\documentclass[ pdflatex,sn-mathphys]{sn-jnl}

\jyear{2021}%

\usepackage{soul,color} 
\usepackage{boldline,multirow}
\begin{document}
  
\title[DistNet2D]{DistNet2D: Leveraging long-range temporal information for efficient segmentation and tracking}


\author*[1]{\fnm{Jean} \sur{Ollion}}\email{jean.ollion@polytechnique.org}
\author[2]{\fnm{Martin} \sur{Maliet}}
\author[3]{\fnm{Caroline} \sur{Giuglaris}}
\author[3]{\fnm{Élise} \sur{Vacher}}
\author*[2]{\fnm{Maxime} \sur{Deforet}}\email{maxime.deforet@sorbonne-universite.fr}

\affil[1]{\orgdiv{SABIlab}, \orgaddress{\city{Die}, \postcode{26150}, \country{France}}}
\affil[2]{\orgdiv{Laboratoire Jean Perrin}, \orgname{Sorbonne Université, Centre National de la Recherche Scientifique}, \orgaddress{\city{Paris}, \postcode{75005}, \country{France}}}
\affil[3]{\orgdiv{Laboratoire PhysicoChimie Curie UMR168}, \orgname{Institut Curie, Paris Sciences et Lettres, Centre National de la Recherche Scientifique, Sorbonne Université}, \orgaddress{\city{Paris}, \postcode{75005}, \country{France}}}

\abstract{
Extracting long tracks and lineages from videomicroscopy requires an extremely low error rate, which is challenging on complex datasets of dense or deforming cells. Leveraging temporal context is key to overcoming this challenge. We propose \text{DistNet2D}, a new deep neural network (DNN) architecture for 2D cell segmentation and tracking that leverages both mid- and long-term temporal information. \text{DistNet2D} considers seven frames at the input and uses a post-processing procedure that exploits information from the entire video to correct segmentation errors. \text{DistNet2D} outperforms two recent methods on two experimental datasets, one containing densely packed bacterial cells and the other containing eukaryotic cells. It is integrated into an ImageJ-based graphical user interface for 2D data visualization, curation, and training. Finally, we demonstrate the performance of \text{DistNet2D} on correlating the size and shape of cells with their transport properties over large statistics, for both bacterial and eukaryotic cells.}


\maketitle
\section{Introduction}
Automated image analysis is revolutionizing the study of cells, enabling scientists to measure their position, velocity, shape, and fluorescent signal intensity. When cells are motile, cell segmentation (localizing cell boundaries) and tracking (associating cells in consecutive images) are combined to follow cell properties over time \cite{boquet2021bioimage}. This data has already revealed fundamental mechanisms in cell motion, especially \textit{in vitro} \cite{friedl2009collective, paul2017cancer, sengupta2021principles, wadhwa2022bacterial}. 
Further improving performance of cell segmentation and tracking will enable us to understand long-term correlation and extend analysis to more and more complex systems. However, extracting long lineages demands an extremely low error rate, which can be difficult to achieve with complex datasets.

Challenges commonly found in biological images stem from: (i) Cell morphology: The shape and size of cells can differ from one another or be very similar within a cell population, and this can change over time, making cell morphology unreliable as a means to segment and track cells accurately. (ii) Cell boundaries: The boundaries between cells can sometimes be difficult to distinguish, as they often touch each other. (iii) Cell movement: The movement of cells can be drastically different from one to the next, or in time, for instance with abrupt changes of direction of motion. (iv) Cellular events: Cellular events, such as mitosis or apoptosis, alter lineages and prove difficult to track using conventional general-object tracking methods. 
For such complex datasets, improving the analysis of individual images is insufficient to overcome the challenges of cell segmentation and tracking. To increase performance, it is necessary to leverage temporal context by considering multiple frames simultaneously.

\subsection{Cell Segmentation}
The main challenge in cell segmentation is not only foreground-background classification, but also the separation of adjacent cells. The emergence of deep neural networks (DNN) has greatly improved the efficiency and robustness of segmentation methods. 
Early DNN-based methods were mostly based on pixel-wise classification \cite{ronneberger2015u}, which resulted in sub-efficient separation of adjacent cells. 
Another popular family of methods involve two sequential DNNs: a region proposal network that generates axis-aligned bounding box candidates followed by a classification network that filters and classifies them \cite{ren2015faster}. Although this type of method is very efficient on natural images, it is complex to train and has been shown to be unable to cope with certain cellular shapes \cite{schmidt2018cell}. 
Other methods do not directly segment cell instances but predict (in this article, the output computed by a DNN is referred to as prediction) pixel-wise features as proxy for segmentation that are subsequently fed to a clustering algorithm: \cite{payer2019segmenting} proposed multi-dimensional embedding with a loss function that pushed dissimilarity between neighbors, \cite{naylor2018segmentation} proposed to predict the Euclidean distance map (EDM), fed to a watershed algorithm. Compared to a binary probability map, EDM has the advantage of emphasizing the boundary between touching cells while being independent of morphology. A popular method that is similar (but not equivalent) predicts radial distance between center and boundaries at predefined angles, which limits the application to convex objects \cite{schmidt2018cell}. \cite{neven2019instance} proposed an efficient method that jointly predicts, for each cell pixel, an offset vector pointing to the cell center and a clustering bandwidth. Similarly, \cite{stringer2021cellpose} predicts a normalized offset vector pointing to the cell center. In case of filamentous bacteria, this method tends to produce over-segmentation. This problem was reduced in \cite{cutler2022omnipose} by predicting the EDM along with an offset vector pointing to the cell medial axis (skeleton), defined by the local maxima of the EDM. 

\subsection{Cell Tracking}
The most straightforward approach to cell segmentation and tracking runs in two independent successive steps: object detection followed by temporal association of detected objects. The recent method DeLTA 2.0 \cite{o2022delta} uses, for the tracking step, a classification neural network to predict the next cell for each cell. However, because predictions are made independently for each cell, this method is likely to produce inconsistent results.
A two-step approach can allow long-term temporal information to feed the tracking algorithm, as segmentation enables data compaction. Notably, \cite{ben2022graph} uses graph neural network to model the entire time-lapse sequence, resulting in very effective tracking. 
The main drawback of the two-step approach is that it is directly limited by the accuracy of the segmentation step. In difficult cases such as high density of similar cells, even a trained expert requires temporal context to perform accurate segmentation. 

\subsection{Combined Segmentation and Tracking}
Temporal information can be leveraged by combining segmentation and tracking into a single operation.
Several recent methods simultaneously train a bounding box detector with a tracker that associates bounding boxes candidates between successive frames \cite{voigtlaender2020siam, chen2021celltrack}. In the context of cell tracking, one limitation of this kind of method is that they have restricted access to the spatial context around the bounding boxes (such as the position of neighboring objects), which is crucial when cells have similar aspects.
An emerging trend is the prediction of the displacement vector of each cell between two successive frames as a proxy for tracking \cite{Hayashida2020MPM, ollion2020distnet, loeffler2022, hayashida2022consistent}, along with a proxy for segmentation or detection. The actual association of cells is performed in a post-processing step. One advantage of this strategy is that it enables simultaneous segmentation and tracking of all cells present in a time window using a single DNN, which likely yields more consistent results for both tasks. It is noteworthy that \cite{Hayashida2020MPM, hayashida2022consistent} do not segment cells but only detect their centers. \cite{ollion2020distnet} introduced an attention layer \cite{vaswani2017attention} in the neural network, and showed that it captures long-range spatial information, in the one-dimensional case of bacterial cells growing in a microfluidic device. \cite{loeffler2022} use a DNN architecture that performs segmentation independently for each frame and thus cannot leverage temporal context for segmentation.
However, several works have shown that performing joint segmentation and tracking improves segmentation by leveraging sequential information \cite{payer2019segmenting, ollion2020distnet}. 
Due to memory limitations, these methods can only use a small temporal neighborhood, e.g. \cite{Hayashida2020MPM, ollion2020distnet, loeffler2022} use pairs of successive frames (t,t+1). More recently, \cite{hayashida2022consistent} have shown that tracking and detection performance can be improved by using a larger temporal neighborhood of six frames as well as a carefully designed loss function that penalizes inconsistencies between detection and tracking.

\vspace{1\baselineskip}
In this work, we describe \text{DistNet2D}, a novel 2D cell segmentation and tracking method, with a carefully designed DNN architecture that leverages mid- and long-term temporal context for both tasks.
Mid-term temporal context is incorporated at the input of the DNN: our method typically considers a 15-frame time window, but this size is adaptable to the features of the dataset and can be much wider if needed. 
Long-term temporal context is incorporated through a post-processing procedure that uses information from the whole video to correct segmentation errors. 
We compare \text{DistNet2D} to two recent methods (DeLTA 2.0 \cite{o2022delta}, EmbedTrack \cite{loeffler2022}) on two experimental datasets that we publish along with this work: a dataset containing phase contrast images of dense communities of motile bacterial cells, and a dataset of fluorescence images of adherent migrating eukaryotic cells.
We also adapted the graphical user interface of BACMMAN software \cite{ollion2019high} for 2D data visualization, curation, and training.
Finally, we demonstrate how \text{DiSTNet2D}'s performance enables us to correlate the size and shape of cells with their motion properties over large statistics, for both bacterial and eukaryotic cells.

\section{Results}
\begin{figure}
\centering
\includegraphics[width=\textwidth]{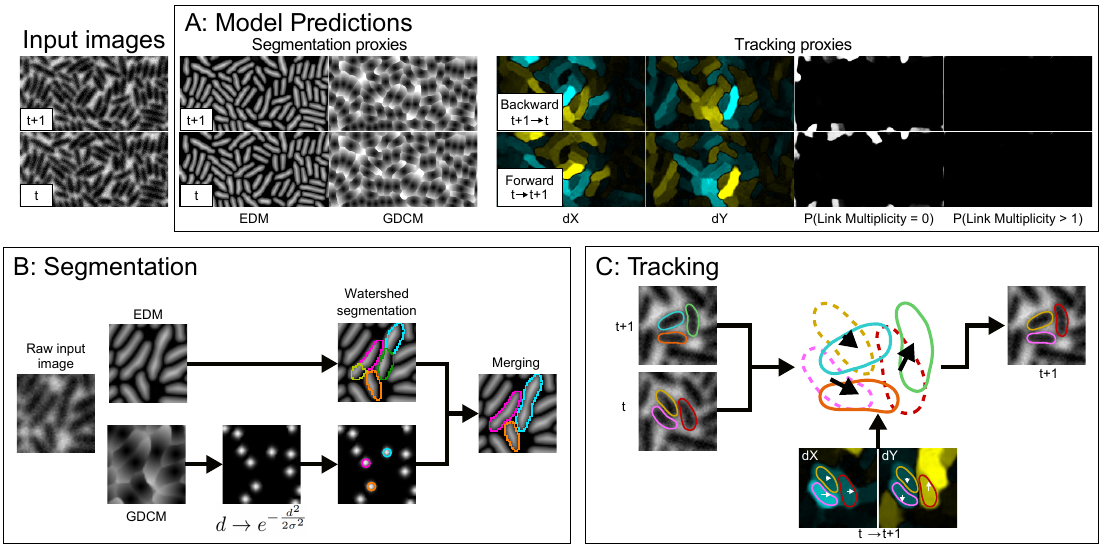}
\caption{Method overview. A: Output of the model for a given frame pair. For each frame, \textit{EDM} is the map of the Euclidean distance to the edge of each cell, \textit{GDCM} is the map of the geodesic distance to the center. For each pair of frames, in each direction, \textit{dX} and \textit{dY} are the cell center displacements from previous frame for each axis, $P(\textit{Link multiplicity}=0)$ and $P(\textit{Link multiplicity}>1)$ are the probabilities that the link multiplicity is zero (no linked cell in the other frame) and strictly greater than one (several linked cells in the other frame), respectively. In this dataset, $P(\textit{Link multiplicity}>1)$ is null because it contains no mitosis or merging cells.
Note that only one frame pair ($t$, $t+1$) is represented, but the model inputs a larger temporal neighborhood and predicts these maps for more frame pairs. The output includes both forward ($t \rightarrow t+1$) and backward ($t+1 \rightarrow t$) predictions. B: Segmentation procedure: A watershed transform is applied to the EDM using regional maxima as seeds, which likely produces over-segmentation. Gaussian function is applied to the predicted GDCM and a watershed algorithm on the Laplacian transform is used to detect centers, which are used to reduce over-segmentation (see main text for details). C: Tracking procedure: The centers of each cell at $t$ are shifted by the predicted displacement between $t$ and $t+1$ (dX and dY). Each cell at $t$ is associated with the cell at $t+1$ in which the shifted center falls. Images are from dataset PhC-C2DH-PA14. Method overview for dataset Fluo-C2DH-HBEC is available in Figure \ref{si_fig:method}.}
\label{fig:method}
\end{figure}

Following the work of \cite{ollion2020distnet, hayashida2022consistent, loeffler2022}, we developed a method that performs segmentation and tracking simultaneously with a single DNN. This strategy has several advantages over methods that perform the tasks independently. First, it leverages temporal information for segmentation, improving the accuracy of the results. Second, it is easier to train and use a single DNN than two separate networks.
Our method is based on a novel DNN architecture that incorporates a sequence of operations designed to blend the information gathered from the different input frames, enabling the use of this information for both segmentation and tracking (see details in Online Methods). 
Specifically, several frames are fed to the DNN, which predicts proxies for both segmentation and tracking (Figure \ref{fig:method}A). 
Using a larger time window is expected to increase temporal consistency, but the number of considered frames is limited by GPU memory. We chose to consider seven frames: three frames before and three frames after the current frame. To enable the DNN to use information at a longer time range without exceeding its memory capacity, we distributed the seven frames across a larger range, by spacing them apart (see Figure \ref{si_fig:interval} for a diagram). The gap between considered frames depends on the dataset, in this work we used values of one and three frames (depending on the dataset). This strategy is referred to as temporal subsampling.

\subsection{Segmentation}
The network predicts two complementary proxies for segmentation, the Euclidean distance map (EDM), which aims at identifying the cell shape, and the geodesic distance to the center map (GDCM), which aims at identifying the cell center. These proxies are predicted within each cell and then combined to produce objects' contours.
More formally: let $B$ be the background, $F$ the foreground and $C_j$ the $j^{th}$ cell ($F = \bigcup_{j} C_j$), $c_j$ its center, $d$ the Euclidean distance function and $d_g$ the geodesic distance function (see section \ref{algo_seg}); for each pixel $i$: 
$$
EDM_i = \left\{
    \begin{array}{ll}
        min(d(i, B), d(i, F \setminus C_j)) \text{ if } i \in C_j\\
        -1 \text{ if } i \in B\\
    \end{array}
\right.
$$
$$
GDCM_i = \left\{
    \begin{array}{ll}
        d_g(i, c_j) \text{ if } i \in C_j\\
        0 \text{ if } i \in B\\
    \end{array}
\right.
$$

The medoid center of the cells is used in order to ensure it is contained in the cell, even for non-convex shapes. For simplicity, it will be referred to as center in this work. 

EDM-based segmentation is efficient even on non-convex cell morphologies, such as bent bacterial cells. It is performed by applying a watershed algorithm on EDM. Watershed is naturally limited to positive values (as the background is set to -1) and seeded from regional maxima of the EDM, which can easily produce over-segmentation, especially in long cells or cells with complex shapes that may contain several local maxima. 

To reduce over-segmentation, we combined contours with predicted centers. Centers are segmented by performing a watershed algorithm on the Laplacian transform of the image that results from the Gaussian function applied to GDCM (see section \ref{algo_seg} for details).  
Two segmented regions in contact are merged if either one of them or both do not contain a segmented center, or if the ratio of intensity amplitude of the centers is below a user-defined threshold (see Figure \ref{fig:method}B). 

Moreover, we also observed that predicting a unique center per cell improves EDM prediction especially in distinguishing neighboring cells instances. 

\subsection{Tracking}
\label{tracking}
Tracking is performed using the prediction of the displacement of the cell centers along the X and Y axis that occurs between two frames. The center of each cell at $t$ is shifted by its predicted displacement between $t$ and $t+1$; if the shifted center falls into a segmented cell at $t+1$, then the two cell instances are associated (see Figure \ref{fig:method}C). This is similar to the procedure used in \cite{loeffler2022}.

To assist the tracking procedure and manage more complex cases, we also predict the \textit{link multiplicity} category in both the forward ($t \rightarrow t+1$) and backward ($t+1 \rightarrow t$) directions, accounting for the expected number of links for each cell (Figure \ref{si_fig:multiplicity}). The possible values for forward link multiplicity are: no next cell (the cell will leave the field of view, or will die), one next cell (regular case, or the cell will fuse with another cell), multiple next cells (the cell will divide). The possible values for backward link multiplicity are: no previous cell (the cell has entered the field of view, or just appeared), one previous cell (regular case, or the cell has just divided), multiple previous cells (the cell has just fused). Formally, we predict each time three probability maps: $P(\textit{Link Multiplicity}=0)$, $P(\textit{Link Multiplicity}=1)$, $P(\textit{Link Multiplicity}>1)$, summing to 1. We assign the link multiplicity category to each cell as the multiplicity with the highest median probability within the cell.

Cells that are predicted to have a single next cell are linked by forward tracking using the predicted forward displacement. 
Cells with multiple next cells (either because of over-segmentation at $t+1$ and not at $t$, under-segmentation at $t$ and not at $t+1$, or a predicted division event) remain unlinked after forward tracking. 
Backward tracking is then applied on unlinked cells that are predicted to have a single previous cell, using predicted backward displacement (Figure \ref{si_fig:forwardbackward}).

Forward and backward tracking allow to assign both merge links ---in which several cells at $t$ are associated to a single cell at $t+1$--- and split links ---in which several cells at $t+1$ are associated to a single cell at $t$. 
When merge links or split links are not confirmed by the link multiplicity category, it means they arise from over/under segmentation errors and they will be corrected in the post-processing stage (see section \ref{seg_corr}). 
Identifying merge and split links also enables a finer definition of metrics and a more accurate diagnosis of origin of errors (incorrect segmentation \textit{vs} incorrect linking), see section \ref{metrics} for details.

\subsection{Post-processing: Segmentation correction}
\label{seg_corr}
A set of rules was designed to correct over/under segmentation errors using the tracking information. We especially observed such errors in the PhC-C2DH-PA14 dataset, which consists of high-speed acquisition videos (typically 100 frames-per-second for a few seconds) of motile rod-shaped bacteria that divide typically every hour. The invagination of the cell membrane at the center of the mother cell is the last stage of the cell cycle before the separation of the two daughter cells. In phase contrast imaging, this invagination appears as a bright region within the cell body, which is similar to the bright area that connects two separate cells when they are in contact (Figure \ref{si_fig:invagination}). It is virtually impossible to determine from a single frame whether an object represents a late-dividing long single cell or two adjacent short cells. The DNN architecture already helps avoid such errors by considering a mid-term temporal context (typically fifteen frames). 
 
\begin{figure}
\centering
\includegraphics[width=\textwidth]{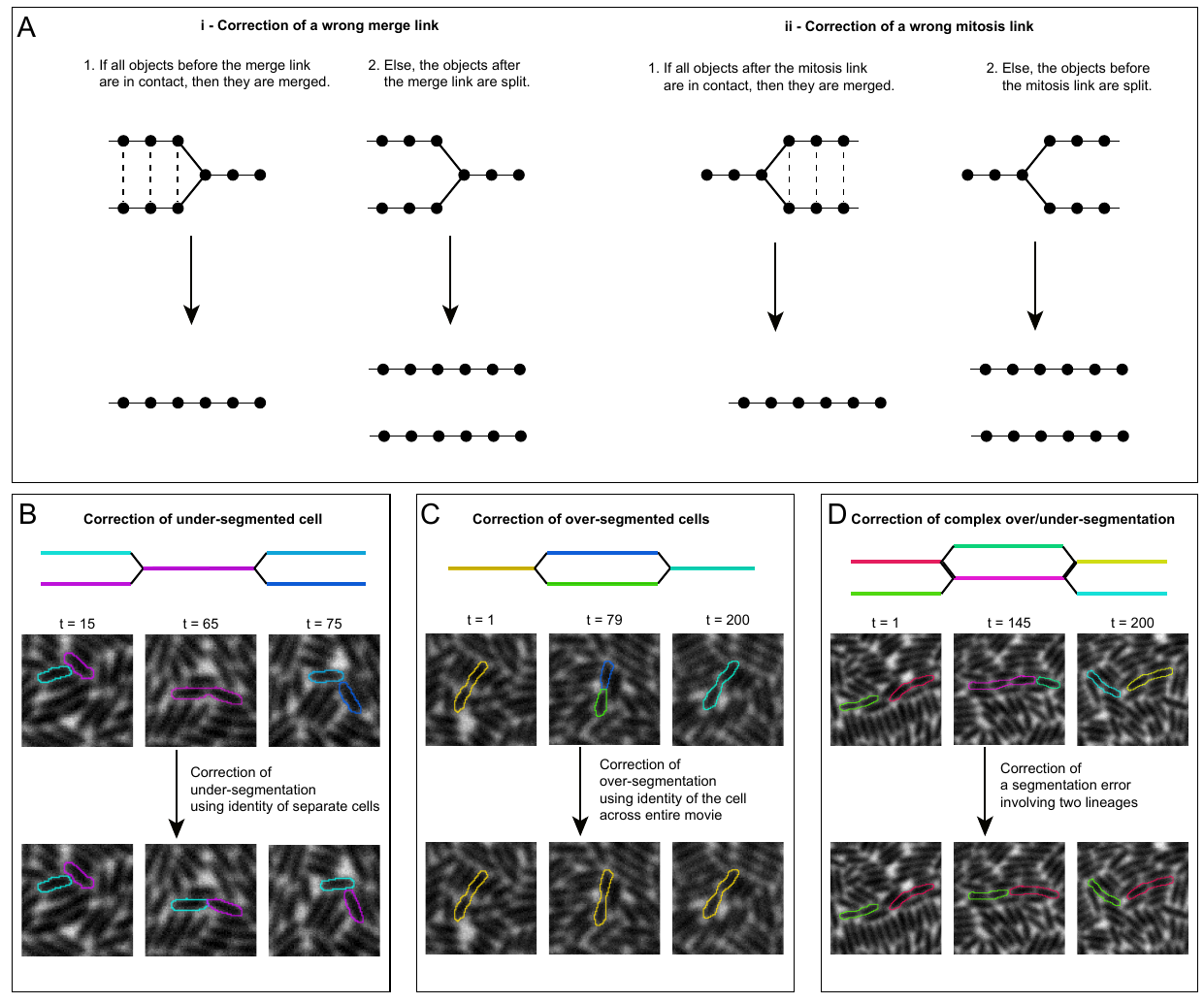}
    \caption{Post-processing uses temporal information on large timescale to correct segmentation errors. A: Diagrams presenting the procedure to correct wrong links. B: Illustrative example of two distinct cells that are transiently detected as one object (under-segmented in frame $t=65$). Because objects involved in the merge link are sometimes seen separated, we can assume the object in frame $t=65$ should be split. C: Illustrative example of one cell that is transiently detected as two objects (over-segmented in frame $t=79$). Because the cell is detected as a single object throughout the entire video (200 frames), except for one frame in which it is seen as two objects, we can assume that the two objects should be merged into one. D: Illustrative example of a more complex error that implies more lineages. In frame $t=145$, one cell is under-segmented and one cell is over-segmented. Here again, temporal information at the scale of the entire video enables us to correct the segmentation errors.}
\label{fig:postprocessing}
\end{figure}

To consider long-term temporal information (which can cover an extended period of time, up to the duration of the entire video), we analyze the trajectories obtained during the tracking step. We focus on the merge and split links. Merge and split links consistent with predicted link multiplicity are considered mitosis or fusion, and are left untouched. In contrast, merge and split links that contradict with predicted link multiplicity are suspected errors.
We treat them using the following general principle: if an object has always been seen as one cell, it should remain as one; however, if it was detected as two distant cells at any point in the past or future, this indicates that it should be considered two cells (Figure \ref{fig:postprocessing}A). This approach is based on the assumption that errors are rare and can be corrected by looking at errorless past and future. The high efficiency of our DNN-based combined segmentation and tracking algorithm supports this assumption.

In practice, for each merge link, we check if all cells before the link are in contact (two objects are considered to be in contact if the distance between their contours is lower than a threshold ; for rod-shaped bacteria, an alignment criterion is also used). If they are, then we merge all of the cells. Otherwise, we split the objects following the link by applying a watershed transform on the EDM (Figure \ref{fig:postprocessing}A-i). Cell fragments are linked to the previous objects using the same procedure as in section \ref{tracking}. If the watershed algorithm generates more fragments than there are previous objects, fragments linked to the same previous object are merged. 
Similarly, for each split link, if all cells after the link are in contact, then we merge all of the cells. Otherwise, we split the objects before the link (Figure \ref{fig:postprocessing}A-ii). Common examples are depicted in Figure \ref{fig:postprocessing}B-D.

\subsection{Model architecture}
\begin{figure}
\centering
\includegraphics[scale=0.65]{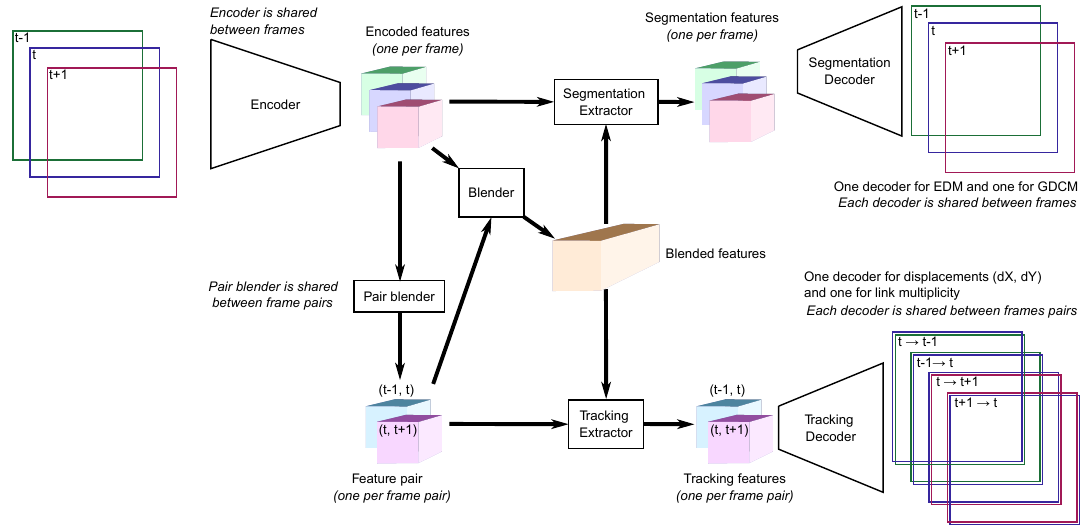}
\caption{Model architecture. The encoder is fed by successive frames (green, blue and red rectangles) and produces encoded features (green, blue and red cubes). Features are processed in pairs (corresponding to successive frames) by the pair blender module, which produces feature pairs. Encoded features and feature pairs are blended together by the blender module (see Figure \ref{si_fig:arch_blender} for details). The segmentation extractor generates three segmentation features corresponding to each frame for both EDM and GDCM, that are decoded by two distinct decoders to produce images of the same size as the input image. Likewise, the tracking extractor generates two tracking features corresponding to each frame pair for both displacement and link multiplicity, that are decoded by two distinct decoders. For simplicity only three frames have been represented but we considered seven frames in this work, and only one segmentation decoder and one tracking decoder are represented instead of two.}
\label{fig:arch}
\end{figure}

The DNN has an encoder-decoder architecture, with a single encoder and one decoder per output type (EDM, GDCM, displacement and link multiplicity). The encoder and decoders are shared between frames for better training efficiency. 
For segmentation outputs (EDM and GDCM), one prediction per frame is made, whereas for tracking outputs (displacement and link multiplicity) one prediction per frame pair is made. For a DNN time window of size $(N-3)\delta+3$ centered on frame $t$, the $N$ considered frames are the three central frames ($t-1$, $t$, $t+1$), and $m=(N-3)/2$ frames on each side of the central frames ($t-1-m\delta$, ..., $t-1-2\delta$, $t-1-\delta$ and $t+1+\delta$, $t+1+2\delta$, ..., $t+1+m\delta$) (Figure \ref{si_fig:interval}). $2N-4$ frame pairs are defined as follow (Figure \ref{si_fig:framepairs}):

\begin{itemize}
    \item $N-1$ frame pairs between consecutive considered frames, for short-range displacements,
    \item $N-3$ frame pairs between the central frame (frame $t$) and each other frame except frames $t-1$ and $t+1$, for mid-range displacements.
\end{itemize}

Figure \ref{fig:arch} displays the global architecture. The detailed architecture of each box is described in Online Methods \ref{si_arch}. Encoder and decoders are mainly composed of residual blocks of two successive convolutions. 
Between the encoder and the decoders, we introduced a blending module ---a sequence of operations that blends the encoded features of all frames together--- and then extract one feature per frame (resp. frame pair) that are fed to segmentation (resp. tracking) decoders. 
Extraction sequences simply contain a distinct convolution per frame (resp. frame pair). Resulting tensors are combined with encoded features (resp. feature pairs). We define the combine operation as a 1x1 convolution applied to the concatenation of two tensors. 

The blending-extraction sequence makes the information from the whole time window available for the prediction of each output at each frame. This contrasts with \cite{loeffler2022} in which segmentation is performed independently on each other frame. 

This architecture has the advantage of allowing proxy predictions to be made only for the central frame and the frame pair (t, t+1) during the prediction phase, which improves speed and reduces memory consumption.

\subsection{Software}
The software associated with our method is BACMMAN \cite{ollion2019high}, an ImageJ \cite{schneider2012nih} plugin that was initially developed for analysis of bacterial cells growing in microchannels, with displacement along the microchannel axis. Such data was naturally displayed on kymographs, in which the horizontal axis represents time. In order to display 2D data, we added the hyperstack visualization mode (see Figure \ref{si_fig:bacmman}), in which lineage information is displayed as colored contours. 
All the features of the graphical user interface of BACMMAN such as interactive navigation through images, manual curation, two-way interplay with R/Python for statistical analysis, are thus available for 2D data. 
BACMMAN was also augmented with new features: generation of training sets as well as \text{DistNet2D} training and prediction can now be performed directly from the software. BACMMAN also provides a command-line interface, enabling its use on a computational cluster.

\subsection{Evaluation Metrics}
\label{metrics}
Objective metrics for segmentation and tracking were previously introduced in \cite{matula2015cell}. In that work, cell tracking results were represented using an acyclic oriented graph, in which nodes corresponded to the detected cells and edges represented links (i.e temporal relations) between them. Metrics were based on the number of operations required to transform the result graph into the reference graph. Those operations were \{split/delete/add\} a node and \{delete/add/change the semantics of\} an edge (e.g.: a change between a split link and a normal link). Correspondence between a reference (R) and a result (S) segmented object were established using the following criterion: $\lvert R \cap S\lvert\:\gt 0.5\cdot \lvert R\lvert$, which implied that each reference cell could correspond to one result cell at most. 
We consider this a limitation because it does not allow to take over-segmentation into account; over-segmented cells were thus systematically considered as false positives. Instead, we used the following criterion: $\lvert R \cap S\lvert\:\gt 0.5\cdot min(\lvert R\lvert, \lvert S \lvert)\;\textit{OR}\;\lvert R \cap S\lvert\:\gt C$ with $C$ a user-defined constant that is typically $50\%$ of the average reference cell size. The second term accounts for cases of under-segmentation and partial overlap with ground truth, where the relative overlap is too low but the absolute overlap is significant. 

This approach identifies four types of segmentation errors: false positives (result cells with no reference counterpart), false negatives (reference cells with no result counterpart), over-segmentation (when $N$ result cells match a given reference cell, $N-1$ over-segmentation are counted) and under-segmentation (when $N$ reference cells match a given result cell, $N-1$ under-segmentation are counted).

We also observed that the procedure proposed in \cite{matula2015cell} does not fully distinguish between tracking and segmentation errors: for instance, over-segmentation of one object into two parts is counted as a false positive segmentation error as well as a false negative link. However, if the over-segmented cell parts were all linked to the correct cell(s), no tracking error should be counted (Figure \ref{fig:metrique}). 
We thus developed a procedure inspired by \cite{matula2015cell} to identify tracking errors that are independent of segmentation errors. In other words, our procedure evaluates the tracking efficiency \textit{per se}, given the segmentation errors. 

To do so, for each frame pair ($t$, $t+1$), we transform the nodes of the result graph so that they match with the nodes of the reference graph by applying four successive operations (Figure \ref{fig:metrique}): splitting under-segmented cells at $t+1$, splitting under-segmented cells at $t$, merging over-segmented cells at $t$ and merging over-segmented cells at $t+1$. At each split/merge operation, links are propagated to the resulting nodes. In the case of splitting, if this implies linking $M$ nodes at $t$ to $N$ nodes at $t+1$, where $M\gt1$ and with $N\gt1$, links are determined by a simple linear assignment algorithm that minimizes the distance between cell centers. In the case of merging, all links are simply added to the resulting node.
After these transformations, all nodes of the transformed result graph correspond to a single node in the reference graph, except for false positives, which have no counterparts in the result graph. 
This enables counting false positive and false negative links: the former are links found in the transformed result graph and not in the reference graph, except links automatically added by splitting an under-segmented object. The latter are links from the reference graph that are not found in the transformed result graph and that do not involve false negative objects. Our choice of link propagation in case of merging during graph transformation would miss some false negative links, that are thus added to the count: in case a result cell was merged at frame $t+1$ but has no link with cells at $t$ and the corresponding reference cell is linked, a false negative link is counted. The same applies for a result cell merged at frame $t$ that has no link with cells at $t+1$.

This procedure is applied for each pair of frames, but this does not tell us how these errors are distributed among the different cell lineages. This information is crucial for evaluating an algorithm, as recovering more error-free cell lineages can be more useful even if it makes more frame-pair-wise errors \cite{mavska2023cell}. We therefore counted the number of error-free cell lineages, allowing for a user-defined tolerance to the frame at which mitosis is detected. 

Lastly, we noticed that many errors arise from cells that are partially out of bounds. We believe that these errors can be easily removed automatically and should not be counted as errors. Therefore, our procedure simply ignores errors that are related to cells that touch edges or lineages that contain at least one cell that touches an edge.

\subsection{Evaluations} \label{section:eval}

\begin{table}
\begin{center}
\caption{Comparison of \text{DistNet2D}, DeLTA 2.0 and EmbedTrack on datasets PhC-C2DH-PA14 and Fluo-C2DH-HBEC. Segmentation errors is the sum of false positive, false negative, under- and over-segmentation, divided by the number of cells in the ground truth. Tracking errors is the sum of false positive and false negative links divided by the number of links in the ground truth. Incomplete lineages is the number of lineages with at least one segmentation or tracking error divided by the number of lineages in the ground truth. For datasets PhC-C2DH-PA14 and Fluo-C2DH-HBEC respectively, the total number of cells is 123,057 and 4,273, the total number of links is 145,306 and 4,890, and the total number of lineages is 561 and 60. As explained in the main text, only cells that do not touch edges (and lineages with no cell touching edges) are taken into account.}\label{table:results}
\resizebox{\textwidth}{!}{%
\begin{tabular}{ l  l  r  r  r  r } 
 \textbf{Dataset} & \textbf{Method} & \textbf{Segmentation} & \textbf{Tracking} & \textbf{Incomplete} & \textbf{Prediction} \\ 
  & & \textbf{errors (\%)} & \textbf{errors (\%)} & \textbf{lineages (\%)} & \textbf{time (s/frame)} \\ \hlineB{2}
  \\
 \multirow{3}{*}{PhC-C2DH-PA14} & DeLTA 2.0 & $1.1$ & $0.22$ & $21$ & $22$ \\ \cline{2-6}
  & EmbedTrack & $1.2$ & $0.14$ & $7.1$ & $2.7$ \\ \cline{2-6}
  & DistNet2D & \underline{$0.24$} & \underline{$0.00$} & \underline{$0.53$} & \underline{$0.51$} \\ 
  \vspace{-2mm} \\
  \hlineB{2}
  \\
 \multirow{3}{*}{Fluo-C2DH-HBEC} & DeLTA 2.0 & $1.9$ & $0.92$ & $37$ & $0.84$ \\ \cline{2-6}
  & EmbedTrack & $0.89$ & $0.51$ & $25$ & $1.4$ \\ \cline{2-6}
  & DistNet2D & \underline{$0.49$} & \underline{$0.10$} & \underline{$5.0$} & \underline{$0.27$} \\
\end{tabular}}
\end{center}
\end{table}

DeLTA 2.0 independently performs segmentation and tracking using two independent U-Net models, with \textit{ad hoc} procedures specifically designed for rod-shaped bacteria, such as a skeletonization of the cell body shape to set a maximal weight at the center of the cell during training.
EmbedTrack simultaneously performs segmentation and tracking using a single DNN, but it uses a total time window of only 2 frames (t,t+1) and only forward predictions. 

\text{DistNet2D} outperforms DeLTA 2.0 and EmbedTrack on all our metrics: segmentation errors, tracking errors, and incomplete lineages (Table \ref{table:results}). Notably, \text{DistNet2D} achieved perfect tracking accuracy on the bacterial dataset. 
Additionally, \text{DistNet2D} ran faster than the other two methods on both datasets.

\newcolumntype{?}{!{\vrule width 1pt}}
\begin{table}
\begin{center}
\caption{Ablation experiments on dataset PhC-C2DH-PA14 (the total number of cells is 123,057, links is 145,306, and lineages is 561). The DNN time window of 2 corresponds to a simplified version of the DNN that considers only frames $(t, t+1)$. The last case, which corresponds to \text{DistNet2D}, has a DNN time window of 15 frames ($N=7$ considered frames and $\delta=3$ using the subsampling definition presented in Figure \ref{si_fig:interval}), included post-processing, and achieved the best performance.}\label{table:ablation}
\begin{tabular}{ p{.14\linewidth} | p{.16\linewidth} ? p{.18\linewidth} | p{.14\linewidth} | p{.17\linewidth}} 
  \multicolumn{2}{c?}{Ablations} & \multicolumn{3}{c}{Results }  \\  \hline
 \textbf{Post-processing} & \textbf{DNN Time window} & \textbf{Segmentation Errors (\%)} & \textbf{Tracking Errors (\%)} & \textbf{Incomplete Lineages (\%)}\\ \hline
 No & 2 & $0.65$ & $0.00$ & $10$ \\\hline
 No & 15 & $0.47$ & $0.00$ & $5.3$ \\\hline
 Yes & 2 & $0.50$ & $0.00$ & $0.89$ \\\hline
 Yes & 15 & $0.24$ & $0.00$ & $0.53$
\end{tabular}
\end{center}

\end{table}

We also evaluated the contribution of several components of our method by performing ablation experiments (Table \ref{table:ablation}).
Switching the DNN time window from two frames (a single frame pair ($t,t+1$), as in EmbedTrack and DeLTA 2.0) to fifteen frames, without post-processing, increased segmentation performance and notably decreased the number of incomplete lineages. 
This shows that mid-term context is leveraged for segmentation and brings temporal consistency. 
While the two-frame version without post-processing exhibits fewer segmentation errors compared to EmbedTrack (0.65\% versus 1.2\%), it also results in more incomplete lineages (10\% versus 7.1\%). This difference arises from the contrasting under-segmentation to over-segmentation ratios observed in the two methods, with under-segmentations having a more pronounced impact on incomplete lineages. 

By leveraging temporal information across the entire video, post-processing effectively corrected suspected errors (defined by merge links and split links that are not confirmed by predicted link multiplicity), leading to a substantial reduction in both object-wise segmentation errors and, more notably, lineage-wise errors, enhancing the overall temporal coherence of predictions.
We acknowledge that post-processing might introduce new errors that will be propagated over entire tracks, but we consider this is an acceptable drawback considering its great performance in terms of reducing incomplete lineages.
We also concede that post-processing works well because the DNN already makes few errors. Post-processing and DNN work synergistically, as shown by the poor performance when both components are turned off.

\begin{figure}
\centering
\includegraphics[width=0.95\textwidth]{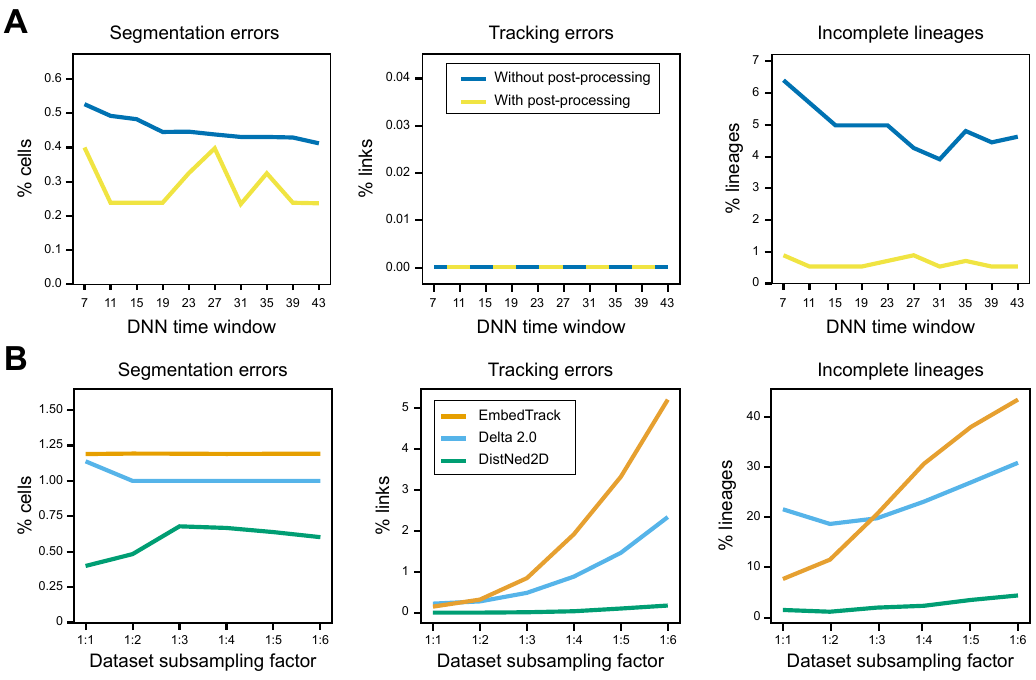}
\caption{Temporal insights of the \text{DistNet2D}'s performance, calculated with dataset PhC-C2DH-PA14.
A: Influence of DNN time window in predictions. DNN time window was varied by changing the gap between considered frames ($\delta$) from 1 to 10, while keeping the number of considered frames constant ($N=7$).
B: Robustness to acquisition subsampling. We compared the three methods (trained with the complete training dataset) on computationally-generated time subsampled versions of the evaluation dataset. Note that for \text{DistNet2D}, $\delta$ was set to 1 for all points, which explains the difference in segmentation errors compared to Table \ref{table:results}, where $\delta=3$.}
\label{fig:sampling}
\end{figure}

We then tested how \text{DistNet2D}'s performances in leveraging long-term information are affected by frame sampling.
First, we varied how the seven considered frames were spread apart, by playing with the parameter $\delta$, which changed the range of the DNN time window (Figure \ref{fig:sampling}A). We found that a wider DNN time window improved the segmentation performance (the tracking efficiency was already very high, even for small $\delta$). This suggests that \text{DistNet2D} benefits from having access to a longer temporal context. The accuracy over entire lineages was also improved with greater $\delta$. 
However, this effect was eliminated after post-processing, possibly because the number of remaining incomplete lineages is too small (lower than 5).

To further assess the influence of time sampling, we compared the performance of \text{DistNet2D}, EmbedTrack, and Delta 2.0 on subsampled evaluation datasets (Figure \ref{fig:sampling}B). While the accuracy of segmentation stayed roughly the same for all methods across the tested range, the accuracy of tracking was sensitive to subsampling, both at the level of individual links or entire tracks. However, \text{DistNet2D} remained fairly accurate for tracking, unlike the other two methods which quickly lost their effectiveness. For example, \text{DistNet2D} performed just as well on the 6-fold subsampled dataset as the other two methods did on the full dataset. Its robustness to subsampling can be explained by the awareness of mid-range temporal information and by the random subsampling performed during data augmentation (see Online Methods \ref{data_aug}).

Overall, the ablation experiments and subsampling experiments confirm that leveraging temporal information improves the segmentation and tracking performance. 

\subsection{Showcase of \text{DistNet2D}'s performance on biological systems}
We demonstrate the potential of \text{DistNet2D} by applying it to bacterial and eukaryotic datasets.
\subsubsection{Monolayer of bacterial cells}
\begin{figure}
\centering
\includegraphics[width=0.95\textwidth]{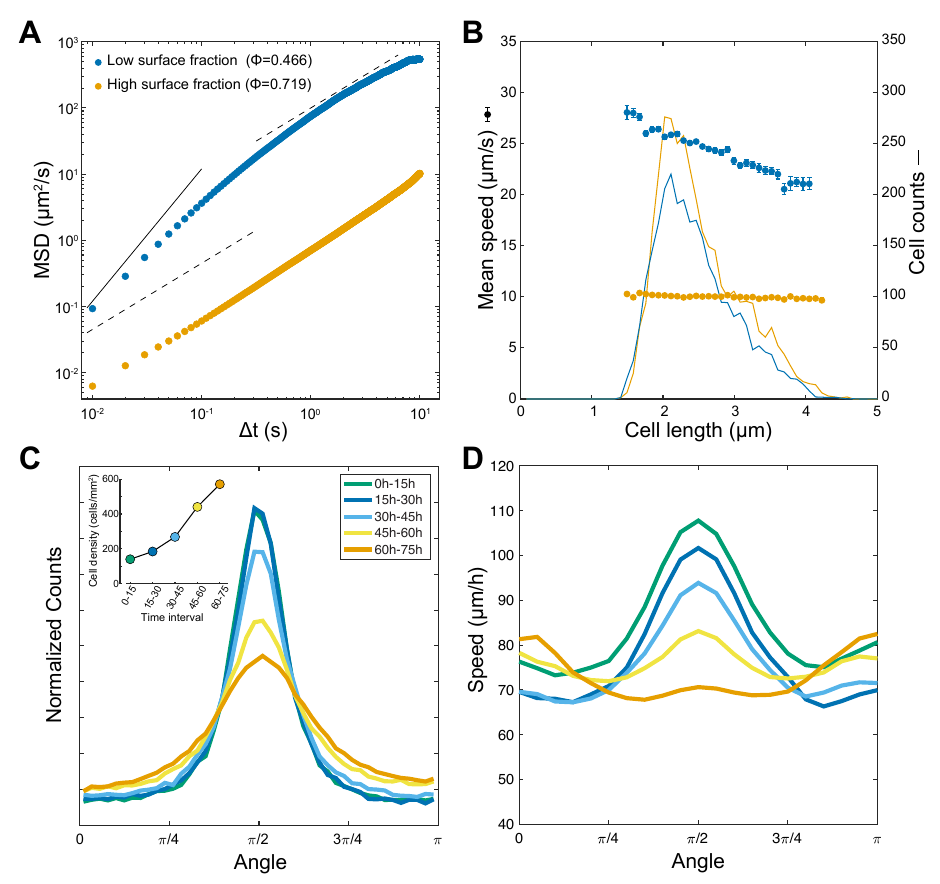}
\caption{A: Mean-squared-displacement measured on bacterial cells at low surface fraction ($\phi=0.466$, blue dots) and high surface fraction ($\phi=0.719$, orange dots). Guide lines have exponent 1 (dashed lines) and 2 (solid line). B: Cell speed as a function of cell length, using the same color code. Error bars are standard errors of the mean (error bars are hidden behind dots for the high surface fraction dataset). Each dot is the mean speed of all cells binned by length, with a bin size of $0.088~\mu m$ (one pixel). Averaging is weighted by the duration of trajectories. The solid lines represent the number of cells in each bin, weighted by the duration of trajectories (a value of 1 is attributed to a cell tracked for 1000 frames).
C: Histogram of angles between the velocity vector and the major axis of the HBEC cell body (obtained by ellipse fitting), for five time intervals (0h-15h, 15h-30h, 30h-45h, 45h-60h, 60h-75h). Each interval includes N=40,939/53,284/76,305/115,860/138,749 data points (respectively). No chirality is measured in the data. Inset: average cell density for each time interval. D: Norm of the velocity vector with respect to the angle between the velocity vector and the major axis of the HBEC cell body, for the same five time intervals.}
\label{fig:dem}
\end{figure}

We measured the mean-squared-displacement (MSD) of \textit{P. aeruginosa} cells at the surface of an agar gel, at a surface fraction of $\phi=0.719$, where the cell monolayer appears "jammed". As shown in section \ref{section:eval}, \text{DistNet2D} was able to extract long tracks: 3404 1000-frame tracks out of an average of 3620 cells that were visible in one field of view. Track duration statistics indicates very few errors (Table \ref{si_table:statistics}). The MSD scales approximately with $t$, confirming diffusive behavior over 3 decades of time (Figure \ref{fig:dem}A). At lower surface fraction ($\phi=0.466$), cells were more motile and a larger fraction of them left the field of view within the duration of the video (1413 1000-frame tracks out of an average of 2911 visible cells). Accordingly, the behavior at short time scales was over-diffusive but remained diffusive at longer time scales. 

We also correlated the length of each cell with its speed. At low density, short cells moved faster than longer cells (Figure \ref{fig:dem}B). We hypothesize that this is a signature of single-cell motility, as viscous drag varies monotonously with the length of a rod \cite{kamdar2022multiflagellarity}. This trend disappeared at high density as cells collectively blocked each other, regardless of their length.

\subsubsection{Eukaryotic cells}
The ability of \text{DistNet2D} to extract precise cell contours and long trajectories enables the correlation of migrating speed and direction with cellular shape. At low density, HBEC cells typically migrate perpendicular to their major axis (Figure \ref{fig:dem}C). At higher density, collisions reorient cells, leading to a less peaked angle distribution. Interestingly, while the largest displacements are perpendicular to the cell body major axis at low density, the trends reverses at high density as some cells move along other cells (Figure \ref{fig:dem}D).

\pagebreak 
\section{Discussion}
\text{DistNet2D} introduces several novel components to address the challenges of segmentation and tracking in bioimages. These components include a novel segmentation method that combines EDM and GDCM, improving the separation of adjacent cells, while being able to segment a wide range of cell morphologies. \text{DistNet2D} also employs a novel approach to predicting backward and forward tracking proxies to handle cell division and fusion events and improve robustness of tracking to under/over-segmentation errors. Long-range temporal context is leveraged in a novel post-processing stage that corrects incorrect merge and split links, relying on the predicted link multiplicity. This stage strongly reduces the lineage-wise error rate by analyzing entire lineages and correcting them when necessary. This strategy is efficient even if errors are also generated at the lineage scale as long as the cell-wise error rate before post-processing is very low.
 
We developed a series of carefully chosen innovations for the implementation of the DistNet2D algorithm.
Tracking and segmentation proxies are predicted by a DNN with a novel architecture designed for leveraging mid-range temporal information for segmentation, while being optimized for size and training efficiency. This range is further increased thanks to a gapped input strategy at no additional GPU memory cost.
The loss function was chosen in order to effectively guide the training process by generating gradients of similar magnitude between segmentation and tracking proxies, regardless of cell size and displacement amplitude. Carefully designed data augmentation allows generalization to diverse imaging conditions without requiring retraining. In particular, we introduced on-the-fly random frame sub-sampling, which improves robustness to changes in acquisition rate, but also increases tracking performances by diversifying displacements during training.
Each of these components plays a crucial role in enhancing the performance of \text{DistNet2D}, making it a powerful tool for analyzing biological processes involving cell movement.

Following \cite{ollion2020distnet}, we tried to introduce an attention layer at the \textit{Pair Blender} module, but this did not improve performance. This is likely because the two datasets used in the study only contained short-range displacements, which could be captured by dilated convolutions. However, an attention layer may be useful for processing datasets with longer-range displacements or displacements that depend on location, such as in microfluidic devices \cite{robert2018mutation,dal2020short}.

We demonstrated the performance of \text{DistNet2D} on two different datasets: a bacterial dataset, where cells are densely packed, have similar shape and only differ in size, and an eukaryotic dataset where cells are sparse, change shape, and divide. Some methods are specifically designed for a precise type of datasets, with \textit{ad hoc} procedures. This is the case for DeLTA 2.0, which was designed for bacterial datasets. Despite its specificity, DeLTA 2.0 under-performs compared to \text{DistNet2D}.
A major strength of \text{DistNet2D} is its ability to leverage both mid- and long-range temporal context, unlike other considered methods: DeLTA 2.0 performs segmentation and tracking separately and does not try to leverage temporal information for segmentation. EmbedTrack performs segmentation and tracking jointly, but its DNN does not blend temporal information and thus does not make it available for all frames, thus segmentation is performed on each frame independently. In both DeLTA 2.0 and EmbedTrack, segmentation does not have access to temporal context. 

We believe \text{DistNet2D} is general and can be used to segment and track any type of moving object in a 2D setting: living or inert, with changing or with fixed shape, with division, with fusion, at any surface density. Any type of imaging can be used: fluorescence, phase contrast, brightfield, etc. The graphical user interface, BACMMAN, is an ImageJ plugin that enables training, curation, manual correction, re-training, distribution of the trained DNN weights to other labs, and data export as tabular data or as label images. Moreover, BACMMAN is able to handle multiple classes of object simultaneously, for instance cell membrane/cell nucleus, cells/foci, head/tail. This could be of particular interest in microbiology, cell biology, soft matter, active matter, but also ethology. The expansion of DistNet2D to 3D datasets is a promising avenue for future exploration. However, the increased memory requirements associated with processing 3D images necessitate further research to ensure efficient training.

Like any other supervised DNN-based method, \text{DistNet2D} training requires a training dataset that can be cumbersome to generate. To facilitate \textit{de novo} dataset creation, BACMMAN includes a DNN-based segmentation method with very low annotation requirements \cite{ludvikova2023near}. Automated tracking can be done in BACMMAN, that includes several tracking methods, such those in Trackmate \cite{tinevez2017trackmate}, an ImageJ plugin that BACMMAN is connected to. The entire pipeline for creating the training dataset (annotations, DNN-based segmentation, manual correction of segmentation, tracking, manual correction of tracking), as well as training of \text{DistNet2D}, can be performed within BACMMAN.

\text{DistNet2D}, a novel method for segmentation and tracking of cells in bioimages, effectively addresses the challenges of segmentation and tracking by leveraging temporal context and employing carefully designed deep learning architectures. The integrated graphical user interface, BACMMAN, offers a comprehensive pipeline that streamlines the process of generating training datasets, training DNN, and applying DNN-based segmentation and tracking, making it more accessible and practical for a broad range researchers. Together, \text{DistNet2D} and BACMMAN form a versatile framework for analyzing biological processes involving cell movement.

\bmhead{Acknowledgments}
This project has received financial support from the ANR X-BACAMAT project (ANR-21-CE30-0025). We are grateful to Pascal Silberzan and Charles Ollion for helpful discussions.

\section*{Declarations}
JO and MD were involved in conceptualization, planning and supervised the work. JO developed and implemented \text{DistNet2D}. MM and MD generated the bacterial datasets. EV and CG generated the eukaryotic datasets. MM analyzed the bacterial data and EV analyzed the eukaryotic data. MM, MD, and JO participated in the manual correction of training datasets. JO and MM performed training, evaluations, and ablation experiments. All authors wrote the manuscript. JO is the founder and director of the company SABILab. Other authors declare no competing interest. Code will be available upon publication.
This work is licensed under a Creative Commons Attribution 4.0 International License (CC-BY-4.0).

\bibliography{bibliography}

\newpage
\section*{Online Methods}
\subsection{Model Architecture}
\label{si_arch}
Encoder module is a sequence of downsampling operations, convolutions, and residual convolution operations (RConv). A downsampling operation is composed of 2x2 max-pooling operations concatenated with a 2x2-strided convolution. \textit{RConv} operations are composed of two successive convolutions and the second activation is applied on the sum of the input of the block and the output of the second convolution. 
For all convolutions, the activation function is ReLU. 
Decoder module is composed of transpose convolutions for upsampling, \textit{RConv} operations and convolutions. 
Encoder and decoder operations are detailed in Table \ref{si_table:arch}.

Several modules are used to blend the encoded features, which organisation is depicted in Figure \ref{fig:arch}. The \textit{Combine} operation concatenates the inputs and applies a convolution with a kernel size 1.
Frame pairs are generated by the \textit{Pair Blender} module that is a \textit{Combine} operation, with a convolution kernel size of 5 and the same number of filters as the encoded features.
\textit{Feature blending} module is depicted in Figure \ref{si_fig:arch_blender}. It consists of three \textit{Combine} operations applied on encoded features and frame pairs, followed by three \textit{RConv} operations.
The number of filters of all the operation of the \textit{Feature blending} module depends on the number of considered frames, $N$, and is calculated using the following formula: $(F \times N) / 2$, $F$ being the number of filters of the encoded features ($F=128$ in this work).

\subsection{Loss functions}
Link multiplicity was trained using the categorical cross-entropy loss function with balanced class frequencies. 
For EDM, GDCM, and displacement, the Pseudo-Huber loss function ---a smooth approximation of the Huber loss--- was used with a delta parameter of 1 \cite{charbonnier1997deterministic, friedman2009elements}. It combines the advantages of the L2 squared loss and L1 absolute loss by being strongly convex near the target but not sensitive to large values. This results in gradients of comparable magnitude for the three predicted features, regardless of the displacement amplitude, or object size and shape.
For all predicted features except EDM, losses were computed only within cells.  

\subsection{Data augmentation}
\label{data_aug} 
We applied several random intensity and geometrical transformations detailed below, as well as random subsampling to simulate larger displacements. The DNN time window was drawn in $[7,\;67]$ for the PhC-C2DH-PA14 dataset ($N=7$, $\delta$ in $[1,\;16]$), and $[7,\;27]$ for the Fluo-C2DH-HBEC dataset ($N=7$, $\delta$ in $[1,\;6]$). Data augmentation was computed on-the-fly during training.

\subsubsection{Intensity transformations}
We randomly scaled the intensity in order to improve generalization and robustness to the presence of long tails in intensity distribution, which is very common in phase-contrast and fluorescence images. Per-image normalization to range $[0,\;1]$ was done using random values in the range of percentiles $[0.01,\;20]$ for the lower bound and $[80,\;99.9]$ for the upper bound. 
In order to improve robustness to loss-of-focus, we applied a Gaussian blur with a random standard-deviation in $[1, 2]$ pixels for 50\% of the images. In order to simulate variations in signal to noise ratio, we added Gaussian noise with a random level. We also performed a random elastic deformation of the histogram as well as in \cite{o2022delta}. 

\subsubsection{Geometrical transformations}
Images were randomly cropped in tiles, with a +/- 10\% random zoom applied simultaneously with cropping. Tile size was set to $192x192$ for PhC-C2DH-PA14 and $384x384$ for Fluo-C2DH-HBEC. Cropping coordinates were also randomly shifted between frames to simulate stage drift/instability and increase displacement variability. Shift was drawn in $[0,\;10]$ pixels for PhC-C2DH-PA14 and $[0,\;20]$ pixels for Fluo-C2DH-HBEC dataset. We also applied random horizontal and vertical flip, $90^\circ$ rotation as well as elastic deformation \cite{van2022elasticdeform} as in \cite{ronneberger2015u}, except that we set a null deformation at edges in order to avoid edge artifacts. 

\subsection{Datasets}
\subsubsection{Fluo-C2DH-HBEC}
This dataset consists of fluorescence images of actin-GFP labelled HBEC cells migrating and dividing on a glass slide, with a time interval of 5 minutes, and a spatial resolution of 0.908 µm/pixel.
Training set consists of one video of 366 frames of 900x700 pixels containing 18,344 cell observations. 
Evaluation set consists of two videos, the first one with $790\times590$ pixels, 99 frames and 2,568 cell observations, and the second one with $900\times700$ pixels, 35 frames and 2,461 cell observations.

\subsubsection{PhC-C2DH-PA14}
This bacterial dataset is made of phase contrast images of \text{\textit{Pseudomonas}} \text{\textit{aeruginosa}} (strain PA14) cells, arranged as a monolayer on an agar gel, with a time interval of 10 ms, and a spatial resolution of 0.088 µm/pixel, acquired at a frame rate of 100 frames-per-second. At this timescale, cell division can be neglected (division time is typically one hour).
The dataset is composed of a training set of two 200-frames videos, of $700\times950$ pixels and $875\times435$ pixels, containing in total about 920,000 bacteria observations. 
The evaluation set is a 200-frame video of $574 \times 515$ pixels. This dataset contains many cells that are partially included in the video, which results in many irrelevant errors. To reduce those errors, predictions were made on the whole video, but evaluated on a $478 \times 364$ pixels subset.

\subsection{Training}
Models were trained for 1000 epochs of 200 steps using the Adam optimizer \cite{kingma2014adam} with learning rate $2.10^{-4}$. Learning rate was decreased by a factor 2 on plateau of 80 epochs. Mini-batch size was set to 24 for PhC-C2DH-PA14 and 16 for Fluo-C2DH-HBEC.

\subsection{Algorithmic details on segmentation}
\label{algo_seg}
GDCM is the proxy for center segmentation.
Geodesic distance is a measure of distance that takes cell shape into account, it corresponds to the shortest path within the mask of the segmented cell. Ground truth GDCM was computed using fast marching method from \href{https://github.com/scikit-fmm/scikit-fmm}{scikit-fmm}.

Centers are segmented by performing a watershed algorithm on the Laplacian transform of scale $\sigma$ of the image that results from the Gaussian function ($d \rightarrow e^{-\frac{d^2}{2\sigma^2}} / \sqrt{2\pi}\sigma$) applied to the predicted GDCM. 
$\sigma$ is a constant of typically 25\% of the object thickness and is limited in the range $[1,\; 3]$ pixels. 
Segmented centers with aberrant size (outside $[50\%,\;200\%]$ of the expected size) or eccentricity $\gt 0.9$ are filtered out. %

\subsection{Experimental details}
\subsubsection{PA14 growth}
Bacterial cells were grown overnight in LB medium at 37°C with aeration. Agar plates of swarming medium (47 mM Na2HPO 4, 22 mM KH 2PO 4, 8.5 mM NaCl, 1 mM MgSO 4, 0.1 mM CaCl 2, 5 g/L casamino acids (Bacto, BD)) were obtained via addition of agar to reach a 0.5\% mass fraction \cite{deforet2023long}. Overnight bacterial suspension was washed twice in PBS buffer and diluted 1000-fold. \text{2 µL} of the washed suspension were added to inoculate an agar plate, which was then flipped and placed inside a 37°C microbiological incubator overnight.

\subsubsection{PA14 observation}
Swarming plates were placed inside an stage-top 37°C incubation chamber (Okolab) mounted on an IX-81 Olympus inverted microscope. Phase contrast videos of 10 seconds at 100 frames per seconds at 40x magnification were acquired using a Blackfly S camera (FLIR), with a resolution of 0.088 µm/pixel. The field-of-view was cropped from $2448\times2048$ to $1000\times1000$ pixels to reach a frame rate of 100 frames-per-second.
The data for training and evaluation sets were obtained with PA14 wild-type.
The data for validation (Figure \ref{fig:dem}A-B) was obtained with strain PA14 $\Delta$PilA, which is flagellated but non-piliated (gift from Dominique Limoli's laboratory in University of Iowa). This strain yields more stereotypical long-term diffusive behavior in monolayers than the wild-type.

\subsubsection{HBEC growth}
Human bronchial epithelial cells (HBECs) (gift from John Minna’s laboratory in Dallas, TX) were cultured in supplemented keratinocyte serum-free medium with L-glutamine (Keratinocyte-SFM with L-glutamine; Gibco). Cells were maintained at 37°C under 5\% CO2 partial pressure and 95\% relative humidity.

\subsubsection{HBEC observation}
HBECs were seeded at low density in wells of a glass bottom 12-well plate in which adequate substrates (glass slide for training and evaluation dataset, PDMS for validation sets) were previously deposited (approximately 26,200 cells/cm2 i.e., 100 000 cells per well). They were left to incubate for a few hours until they were fully attached. Prior to imaging, cell monolayers were rinsed with PBS and 5 mL of fresh medium was added. Time-lapse multi-field experiments were performed on automated inverted microscopes equipped with thermal and CO2 regulations. Experiments lasted typically several days. The displacements of the sample, the illumination sequences and the camera acquisitions were computer-controlled by Metamorph software (Universal Imaging). Experiments were performed with a 10X objective in fluorescence microscopy. The images of the same field of view were taken every 5 min. We used a Leica DM IRB with camera R600 (Qimaging, field of view of $1344\times1100$ pixels, with a resolution of 0.908 µm/pixel) for training and evaluation sets, and a Olympus IX71 with camera Prime BSI (Teledyne Photometrics, field of view of $1024\times1024$ pixels, with a resolution of 1.3 µm/pixel) for validation (Figure \ref{fig:dem}C-D).

\subsection{Computation}
All computations (training, evaluation, validation) were performed on computers equipped with an NVIDIA RTX A6000 48GB graphics card, 128GB RAM, Intel Xeon W-2255 running under Microsoft Windows 11, or AMD Ryzen 7 5800X running under Ubuntu Linux 22.04. Prediction times given in Table \ref{table:results} were calculated on the latter.
Data analysis for Figure \ref{fig:dem} was done with MATLAB (The Mathworks) from the CSV files extracted from BACMMAN, which contain all relevant cellular data (track ID, frame ID, XY coordinates, major axis length, major axis orientation).

\pagebreak 

\section*{Supplementary tables and figures}

\renewcommand{\thefigure}{S\arabic{figure}}
\renewcommand{\thetable}{S\arabic{table}}
\renewcommand{\thesection}{S\arabic{section}}
\renewcommand{\figurename}{Supplementary Figure}
\renewcommand{\tablename}{Supplementary Table}

\setcounter{figure}{0}
\setcounter{table}{0}
\setcounter{section}{1}

\begin{table}[!ht]
\begin{center}
\caption{Details of operations by layer in encoder and decoder modules. As described in the main text, \textit{Down} stands for downsampling, \textit{RConv} for residual convolution, \textit{Conv} for 2D convolution, \textit{Up} for upsampling and \textit{BN} for batch-normalization. 2D Spatial dropout with 0.2 proportion is added to underlined operations. Numbers in parentheses after the operations are kernel size, number of output filters, and in case of RConv, dilation rate applied to the second convolution. Note that for the decoder associated to displacement, there are two distinct upsampling operations for each axis at layer 1. }
\begin{tabular}{| c | c | p{.5\linewidth} | p{.23\linewidth} |} 
 \hline
 Model & Layer & Encoder & Decoder \\ 
 \hline\hline
 D2 & 1 & Down(3x3, 32) & Up(4x4, 1) \\ 
 \hline
 D2 & 2 & Conv(3x3, 32), Conv(5x5, 32), Down(3x3, 128) & Up(4x4, 32), ResConv(3x3, 32), ResConv(3x3, 32) \\
 \hline
 D2 & 3 & \underline{RConv(5x5, 128, 2)}, \underline{RConv(5x5, 128, 3)}, \underline{RConv(5x5, 128, 4)}, \underline{RConv(5x5, 128, 2)}, \underline{RConv(5x5, 128, 3)}, Conv(5x5, 128), BN & \underline{Conv(3x3, 64)}, \underline{RConv(3x3, 64)}, Conv(3x3, 64), BN\\
 \hline
 D3 & 1 & Down(3x3, 32) & Up(4x4, 1) \\ 
 \hline
 D3 & 2 & Conv(5x5, 32), Down(3x3, 64) & Up(4x4, 32), RConv(3x3, 32), RConv(3x3, 32) \\
 \hline
 D3 & 3 & RConv(3x3, 64), RConv(3x3, 64), Down(3x3, 192) & Up(4x4, 64), RConv(3x3, 64), RConv(3x3, 64) \\
 \hline
 D3 & 4 & \underline{RConv(5x5, 128, 2)}, \underline{RConv(5x5, 128, 3)}, \underline{RConv(5x5, 128, 4)}, \underline{RConv(5x5, 128, 2)}, \underline{RConv(5x5, 128, 3)}, Conv(5x5, 128), BN & \underline{Conv(3x3, 64)}, \underline{RConv(3x3, 64)}, Conv(3x3, 64), BN\\
 \hline
\end{tabular}
\end{center}
\label{si_table:arch}
\end{table}

\begin{table}
    \centering
    \begin{tabular}{ccc}
         & Low density & High density\\ \hline
        Surface fraction & 0.446 & 0.719\\ \hline
        Average number of cells per image & 2911 & 3620\\ \hline
        Number of tracks & 7788 & 3758\\ \hline
        Number of tracks analyzed for Figure \ref{fig:dem}A-B ($>10$ frames) & 7620 & 3675\\ \hline
        Number of complete tracks (1000 frames) & 1413 & 3404\\ \hline
        Number of incomplete tracks ($<1000$ frames) & 6375 & 354\\ \hline
        Number of track ends* & 109 & 5\\ \hline \vspace{5mm}
    \end{tabular}
    \caption{Statistics of track duration. *The last line is the number of track ends that are neither at the last frame nor within a distance of 3 $\mu m$ from the image boundary. These track ends are biologically irrelevant and are therefore considered segmentation errors. Notably, there are only five such cell segmentation errors in the high-density video, out of approximately 1000 frames $\times$ 3620 cells $= 3.6 \cdot 10^6$ cell instances. Manual review confirmed these errors are temporary under-segmentations (case B of Figure \ref{fig:postprocessing}), involving two pairs of cells.}
    \label{si_table:statistics}
\end{table}

\begin{figure}[p!]
\centering
\includegraphics[width=0.95\textwidth]{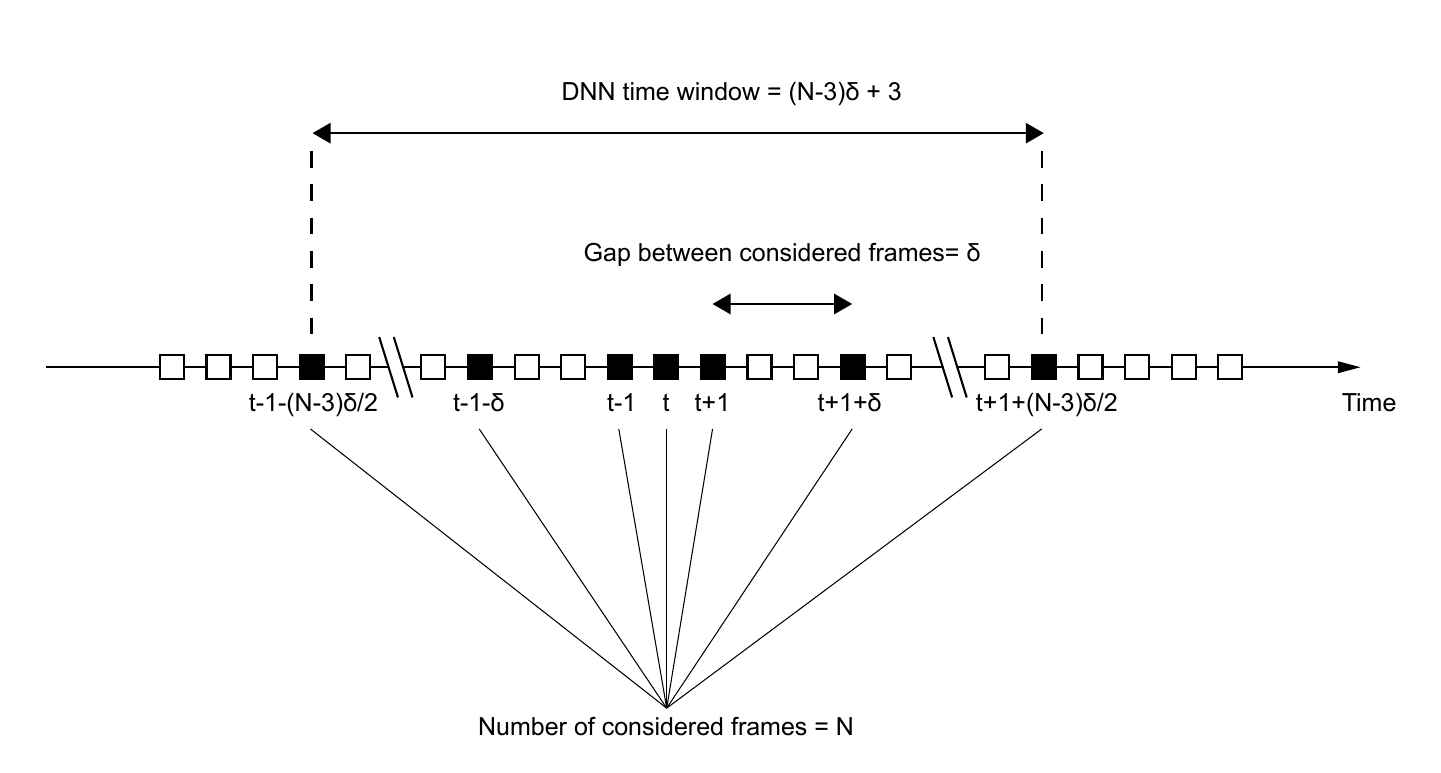}
\caption{Diagram of the subsampling procedure, presenting the definition of number of considered frames ($N$), gap between considered frames ($\delta$), DNN time window ($(N-3)\delta + 3$). Frames immediately before and after the central frame are always included as they are used to compute the displacements.}
\label{si_fig:interval}
\end{figure}

\begin{figure}[p!]
\centering
\includegraphics[width=0.95\textwidth]{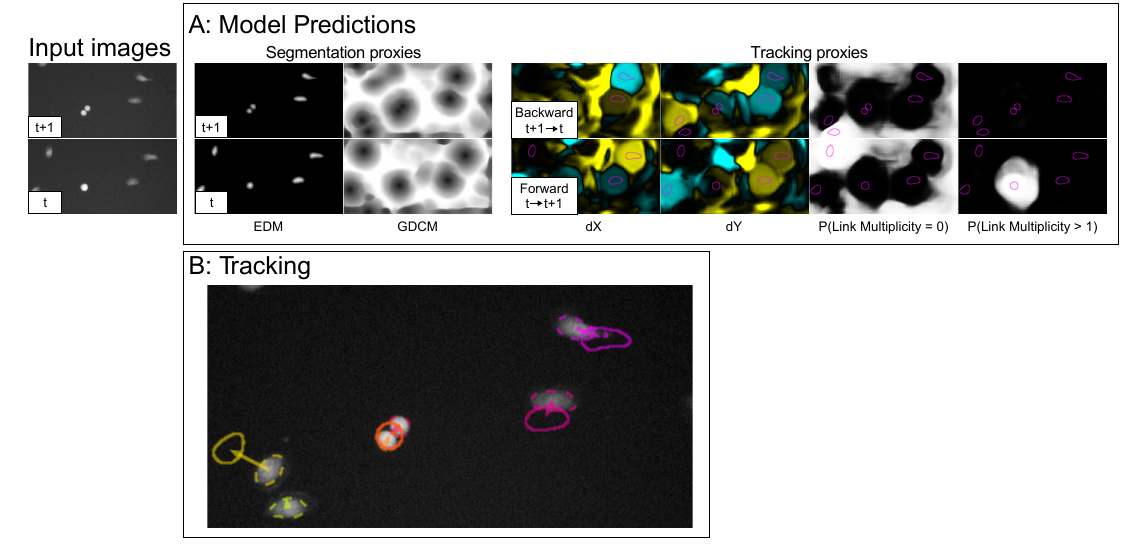}
\caption{Prediction results on Fluo-C2DH-HBEC. A: input and output of the model. \textit{EDM} is the map of the Euclidean distance to the edge of each cell, \textit{GDCM} is the map of the geodesic distance to the center, \textit{dX} and \textit{dY} are the cell center displacement for each axis, above: from $t$ to $t+1$ and below from $t+1$ to $t$. \textit{No link probability} is the probability that a cell has linked cell in the other frame and \textit{multiple link probability} is the probability that a cell has several linked cells in the other frame. For a better readability, contours of segmented cells are represented in the 8 rightmost images. Note that only one frame pair ($t$, $t+1$) is represented, but the model inputs a larger temporal neighborhood and predicts these maps for more frame pairs (see main text). B: Tracking result: contours of cells are represented by a dashed line and contours of assigned previous cells are represented by a solid line. Arrows represent the predicted displacement from $t$ to $t+1$.}
\label{si_fig:method}
\end{figure}

\begin{figure}[p!]
\centering
\includegraphics[width=0.95\textwidth]{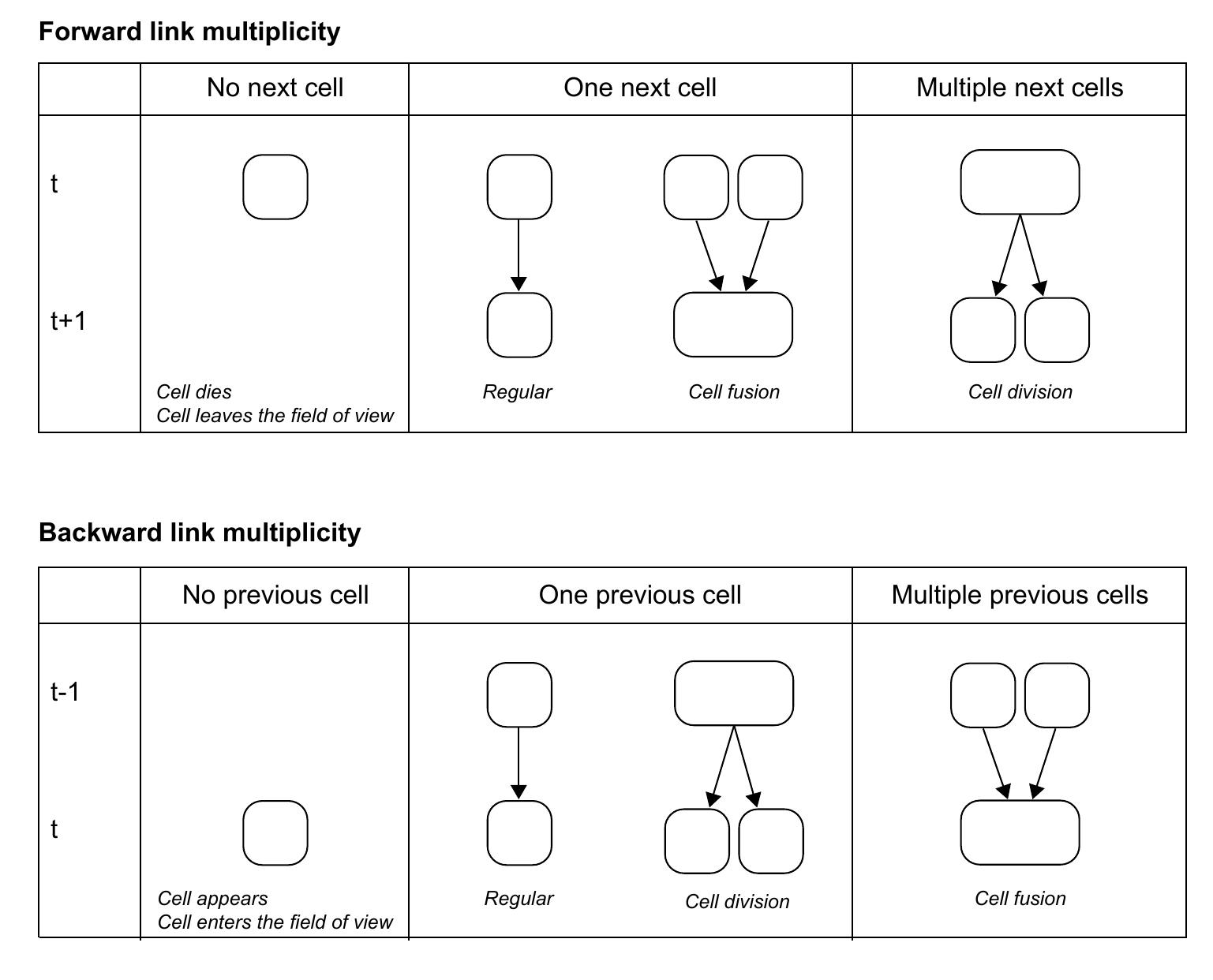}
\caption{Diagram presenting the forward link multiplicity and backward link multiplicity, defined for cells at time $t$.}
\label{si_fig:multiplicity}
\end{figure}

\begin{figure}[p!]
\centering
\includegraphics[width=0.95\textwidth]{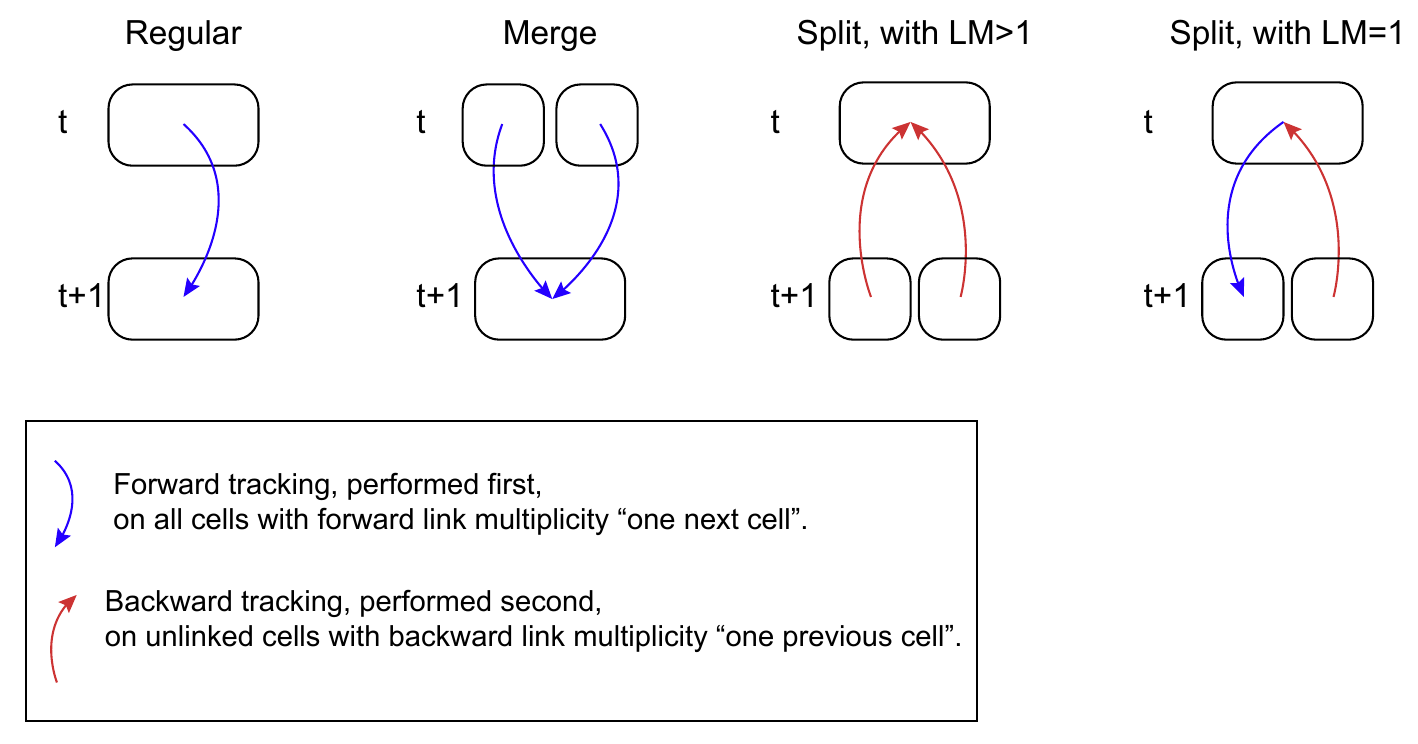}
\caption{Diagram presenting forward and backward tracking. The first case is a single link between two objects. The second case is either an over-segmentation at time $t$, an under-segmentation at time $t+1$, or a predicted merging event. The third case is a predicted split event. The fourth case is either an over-segmentation at time $t+1$, or an under-segmentation at time $t$.}
\label{si_fig:forwardbackward}
\end{figure}

\begin{figure}[p!]
\centering
\includegraphics[width=0.95\textwidth]{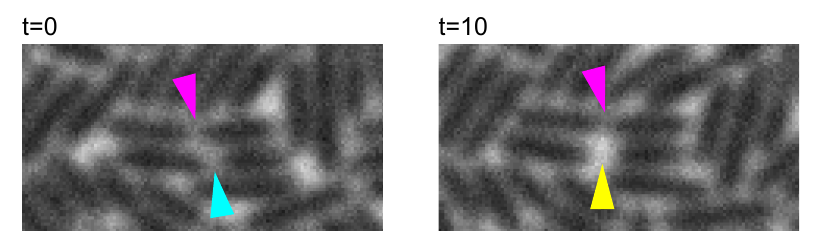}
\caption{Cells exhibit a brighter region at the middle of the body, where the invagination of the cell membrane precedes the complete separation of the two daughter cells (depicted by the magenta arrowheads). The contact between two separate cells look similar (cyan arrowhead). This confusion can only be resolved when the two separate cells actually move apart (yellow arrowhead).}
\label{si_fig:invagination}
\end{figure}

\begin{figure}[p!]
\centering
\includegraphics[width=0.95\textwidth]{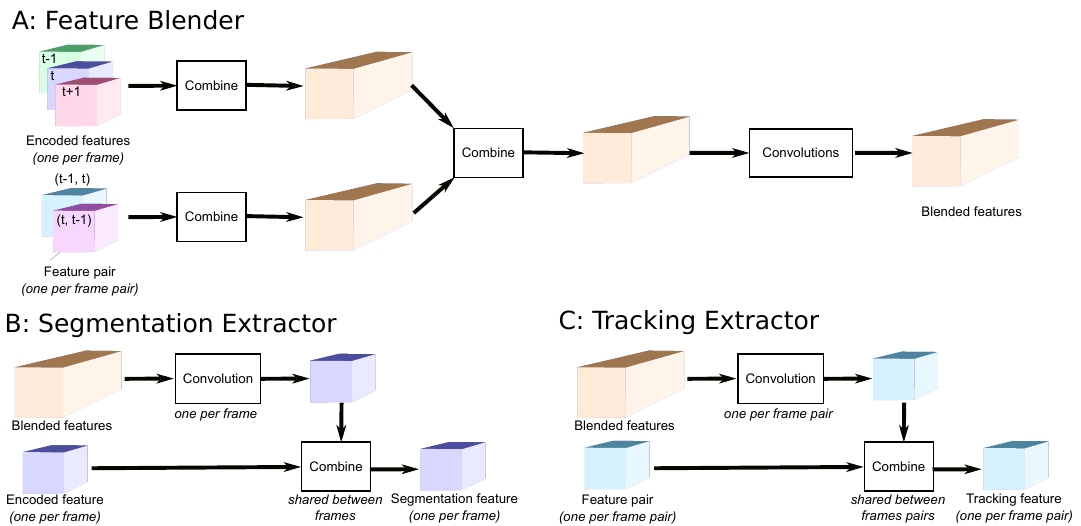}
\caption{Blender module architecture. Encoded features (green, blue and red cubes) and feature pairs (turquoise and magenta cubes) are respectively combined. Combined feature and feature pairs are then combined together. Three residual convolutions are then applied to combined features. For clarity only three frames have been represented but we considered seven frames in this work.}
\label{si_fig:arch_blender}
\end{figure}

\begin{figure}[p!]
\centering
\includegraphics[width=0.95\textwidth]{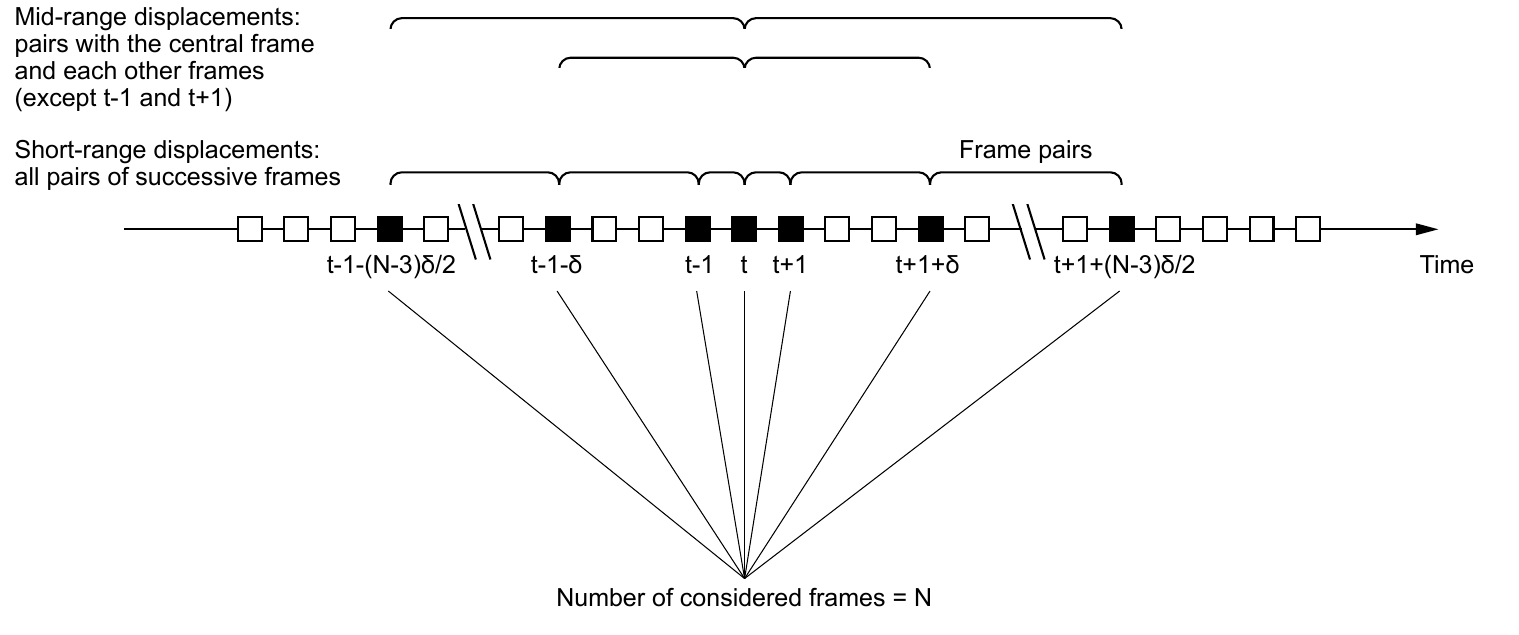}
\caption{Diagram for the definition of the $2N-4$ frame pairs used as input of the DNN.}
\label{si_fig:framepairs}
\end{figure}

\begin{figure}[p!]
\centering
\includegraphics[width=0.95\textwidth]{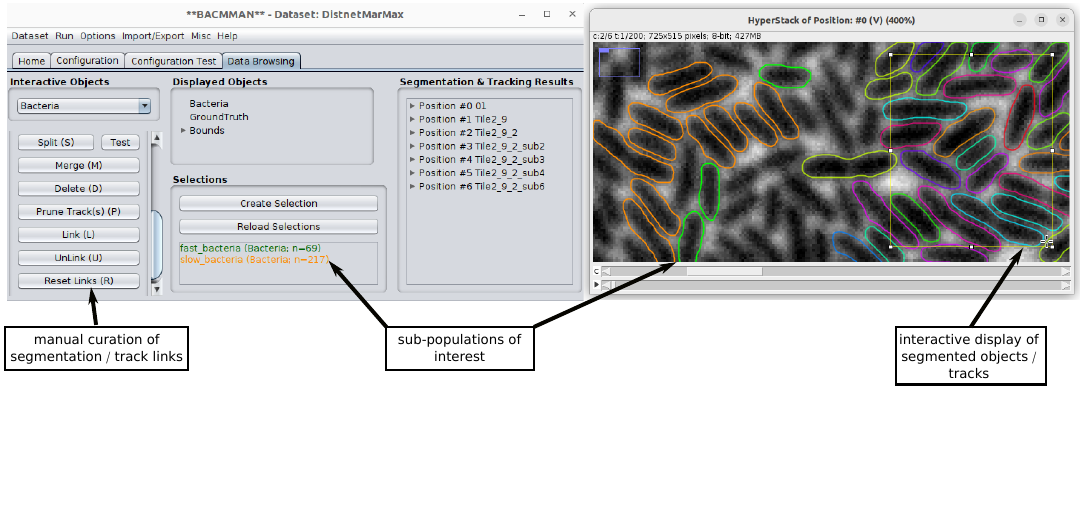}
\caption{Adaptation of BACMMAN's GUI for hyper-stack visualization. Left: example of hyper-stack visualization. On can select objects directly on the image and their tracks are displayed as coloured contours. Left panel displays a part of the GUI of BACMMAN, that enables manual curation of segmentation and tracking, as well as visualization and navigation of sub-populations of interest (selections), displayed here in orange and green. Those selections can be defined from an external statistical processing software such as Python, R or Matlab.}
\label{si_fig:bacmman}
\end{figure}

\begin{figure}[p!]
\centering
\includegraphics[width=0.9\textwidth]{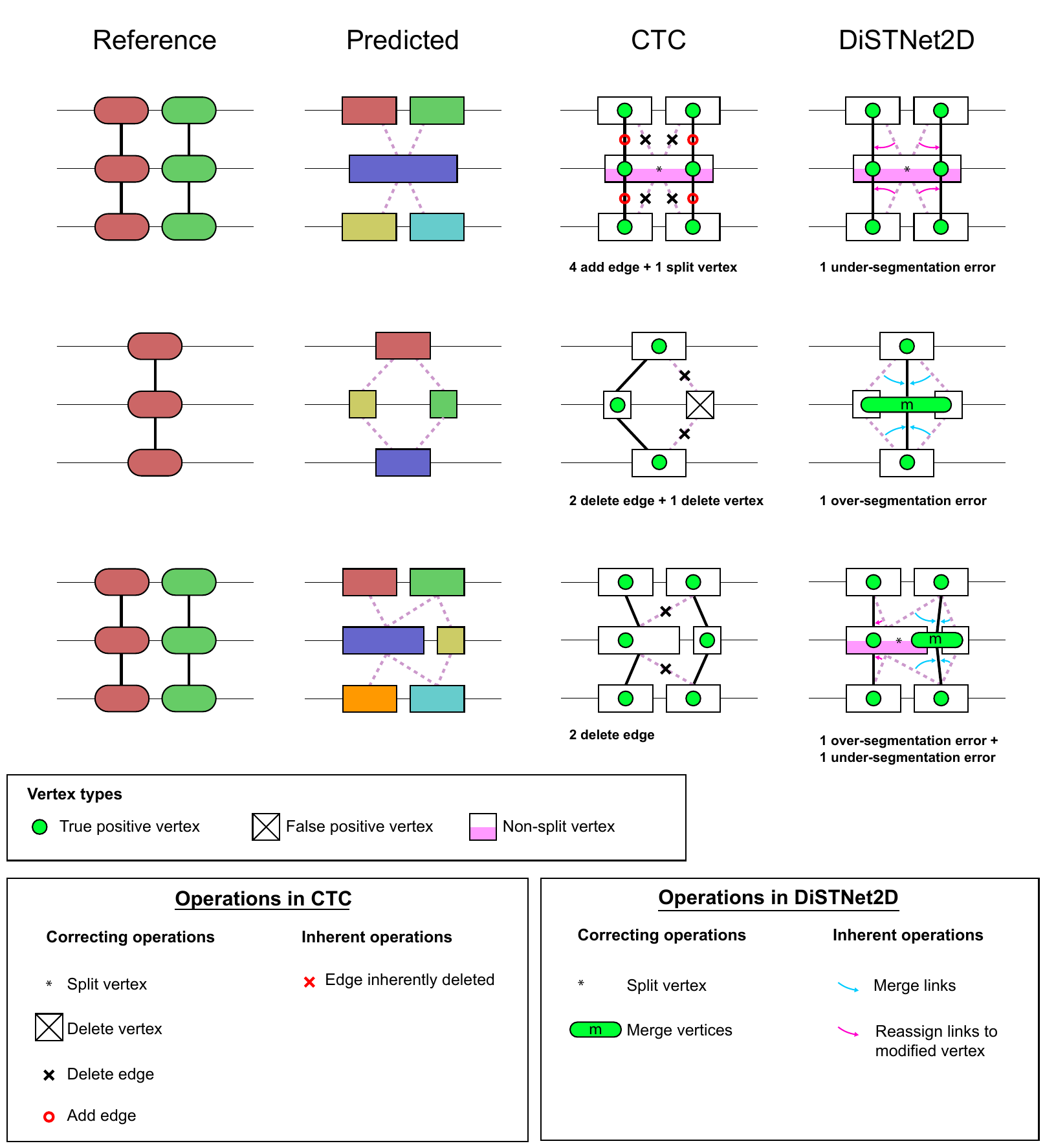}
\caption{CTC vs \text{DistNet2D} metrics. Three examples of segmentation errors (corresponding to examples shown in Figure \ref{fig:postprocessing}) and how these errors are counted in CTC and in \text{DistNet2D}. The first column represents the ground truth. The second column represents the incorrect prediction. The third column represents the error counting according to \cite{matula2015cell}. The last column represents the error counting used in this work. Note that the three examples do not depict all possible error types, but instead highlight how some segmentation errors are also counted as tracking errors in CTC but not in \text{DistNet2D}. 
In the first case (under-segmentation error), we do not count any tracking error: original edges are not deleted at split operation, but modified by minimizing distances.
In the second case (over-segmentation error), we do not count any tracking error if both over-segmented cells are linked to a unique cell at previous and next frames.
In the third case (segmentation errors involving two lineages), CTC allows a reference object to match with only one single predicted object. Under-segmentation is thus not detected. On the contrary, \text{DistNet2D} detects here one under-segmentation, one over-segmentation, and no tracking error.
}
\label{fig:metrique}
\end{figure}

\clearpage
\section*{Supplementary videos}
\renewcommand{\figurename}{Supplementary Video}
\setcounter{figure}{0}

\begin{figure}[h]
\centering
\includegraphics[scale=0.3]{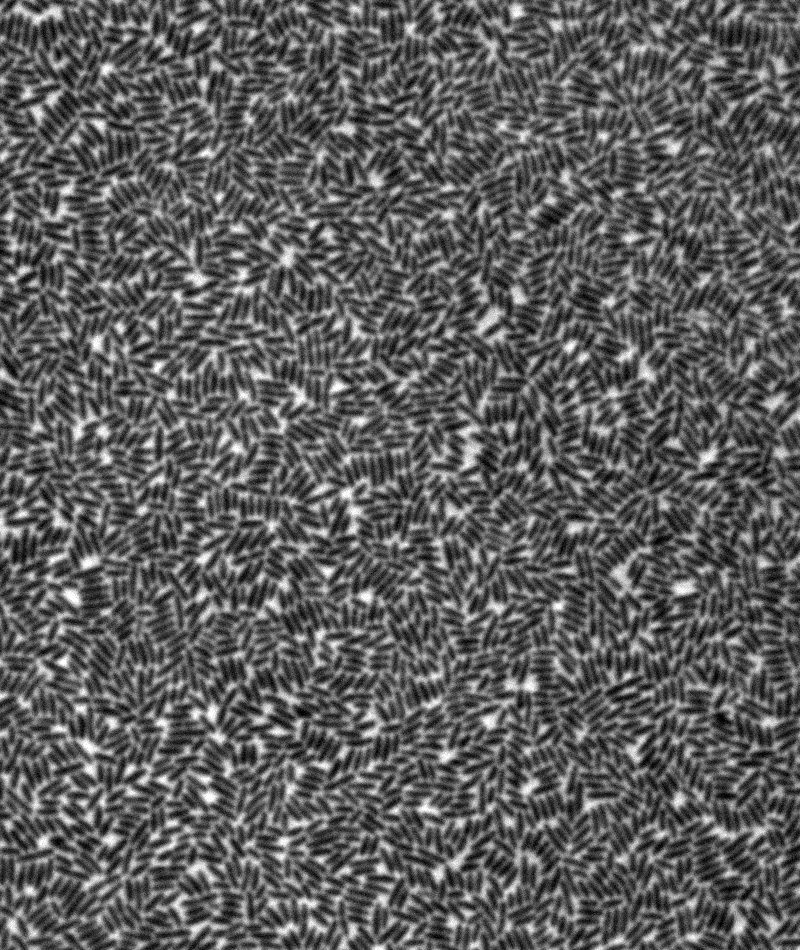}
\caption{Training video \#1 of dataset PhC-C2DH-PA14. To reduce the total video size, the image stack was time-subsampled 2-fold, but the video still plays in real time (50 fps instead of 100 fps).}
\label{si_vid:TrainingPA14_1}
\end{figure}

\begin{figure}[p!]
\centering
\includegraphics[scale=0.3]{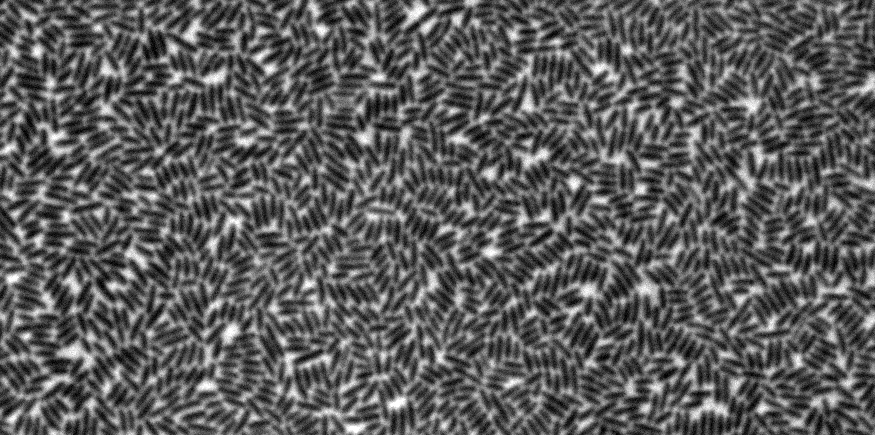}
\caption{Training video \#2 of dataset PhC-C2DH-PA14. To reduce the total video size, the image stack was time-subsampled 2-fold, but the video still plays in real time (50 fps instead of 100 fps).}
\label{si_vid:TrainingPA14_2}
\end{figure}

\begin{figure}[p!]
\centering
\includegraphics[scale=0.3]{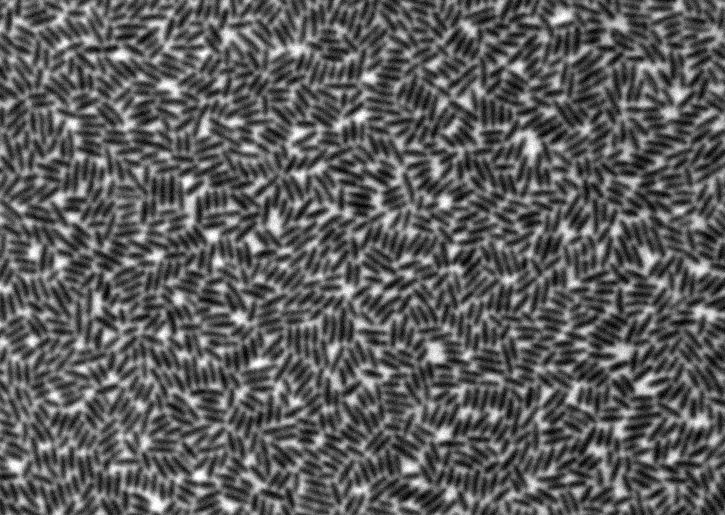}
\caption{Evaluation video of dataset PhC-C2DH-PA14. To reduce the total video size, the image stack was time-subsampled 2-fold, but the video still plays in real time (50 fps instead of 100 fps).}
\label{si_vid:EvaluationPA14}
\end{figure}

\begin{figure}[p!]
\centering
\includegraphics[scale=0.3]{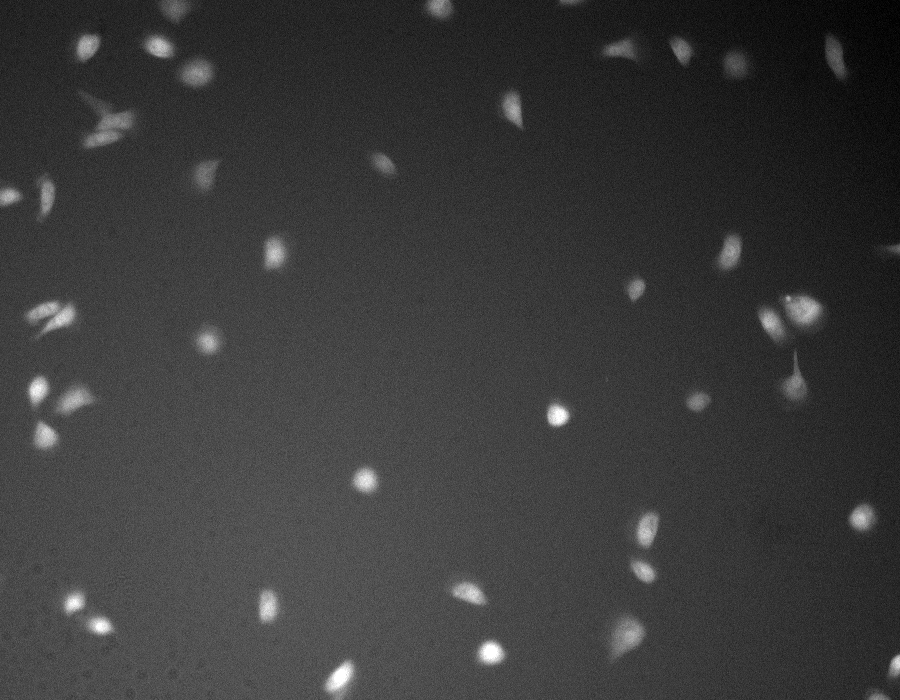}
\caption{Training video of dataset Fluo-C2DH-HBEC.}
\label{si_vid:TrainingHBEC}
\end{figure}
\begin{figure}[p!]
\centering
\includegraphics[scale=0.3]{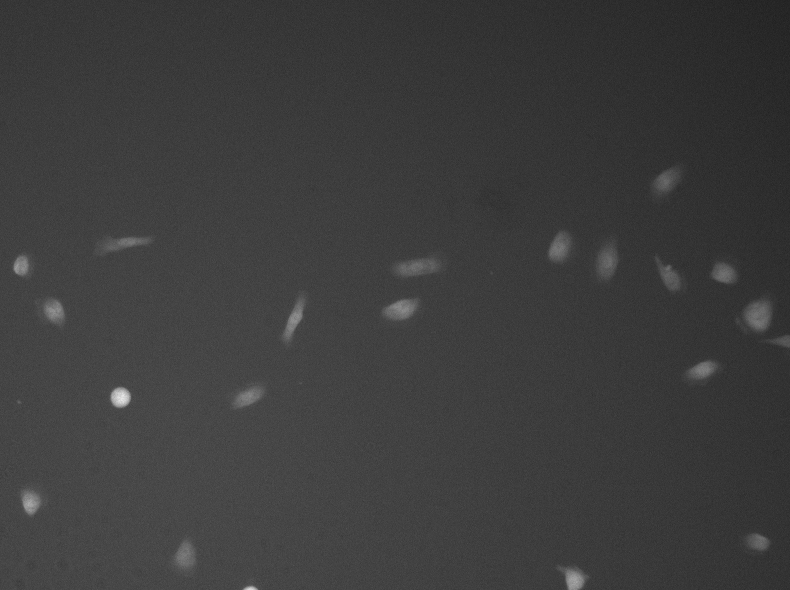}
\caption{Evaluation video \#1 of dataset Fluo-C2DH-HBEC.}
\label{si_vid:TrainingHBEC}
\end{figure}

\begin{figure}[p!]
\centering
\includegraphics[scale=0.3]{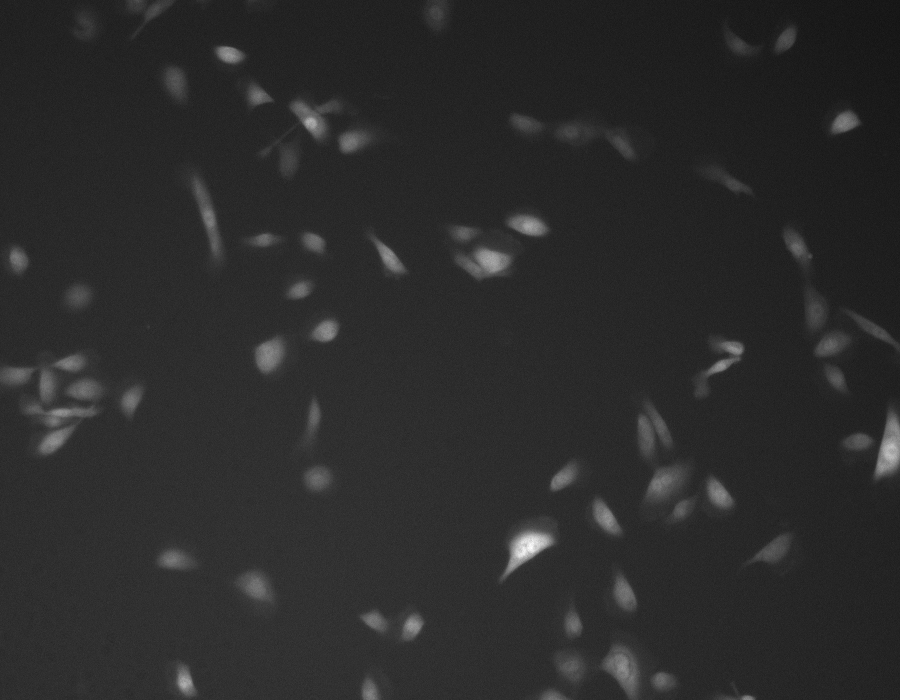}
\caption{Evaluation video \#2 of dataset Fluo-C2DH-HBEC.}
\label{si_vid:TrainingHBEC}
\end{figure}

\begin{figure}[p!]
\centering
\includegraphics[scale=0.3]{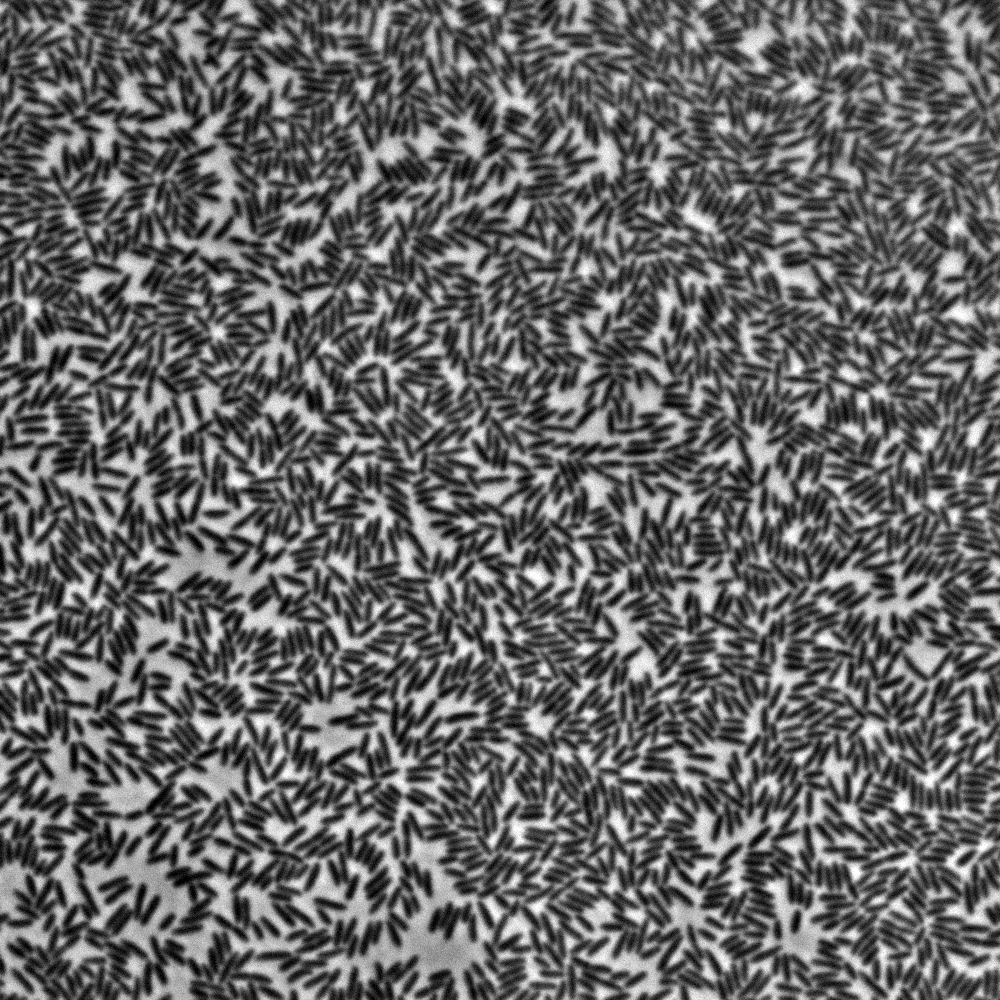}
\caption{Video of the low density monolayer of PA14 $\Delta$PilA cells, used for Figure \ref{fig:dem}A-B. To reduce the total video size, only the first second of the 10-second dataset was included and the image stack was time-subsampled 2-fold. The video still plays in real time, at 50 fps instead of 100 fps.}
\label{si_vid:LowDensityPA14}
\end{figure}

\begin{figure}[p!]
\centering
\includegraphics[scale=0.3]{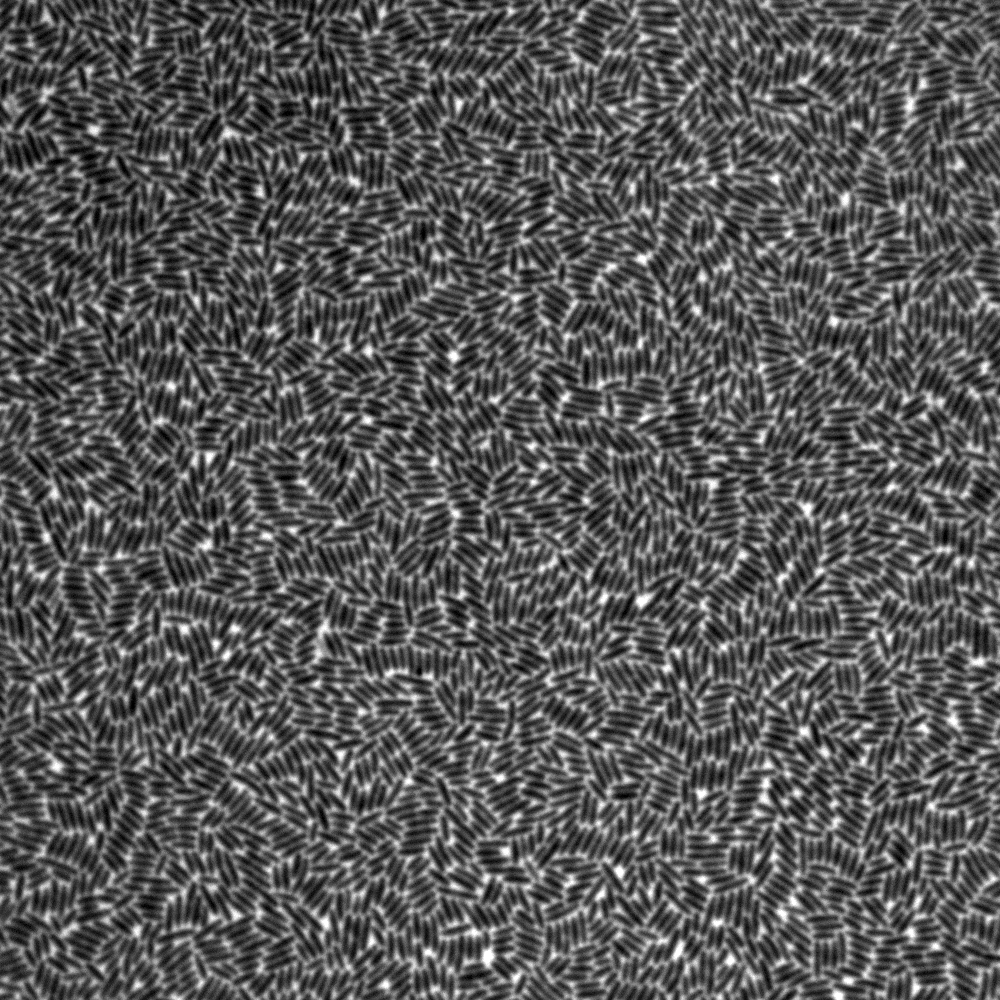}
\caption{Video of the high density monolayer of PA14 $\Delta$PilA cells, used for Figure \ref{fig:dem}A-B. To reduce the total video size, only the first second of the 10-second dataset was included and the image stack was time-subsampled 2-fold. The video still plays in real time, at 50 fps instead of 100 fps.}
\label{si_vid:HighDensityPA14}
\end{figure}

\begin{figure}[p!]
\centering
\includegraphics[scale=0.3]{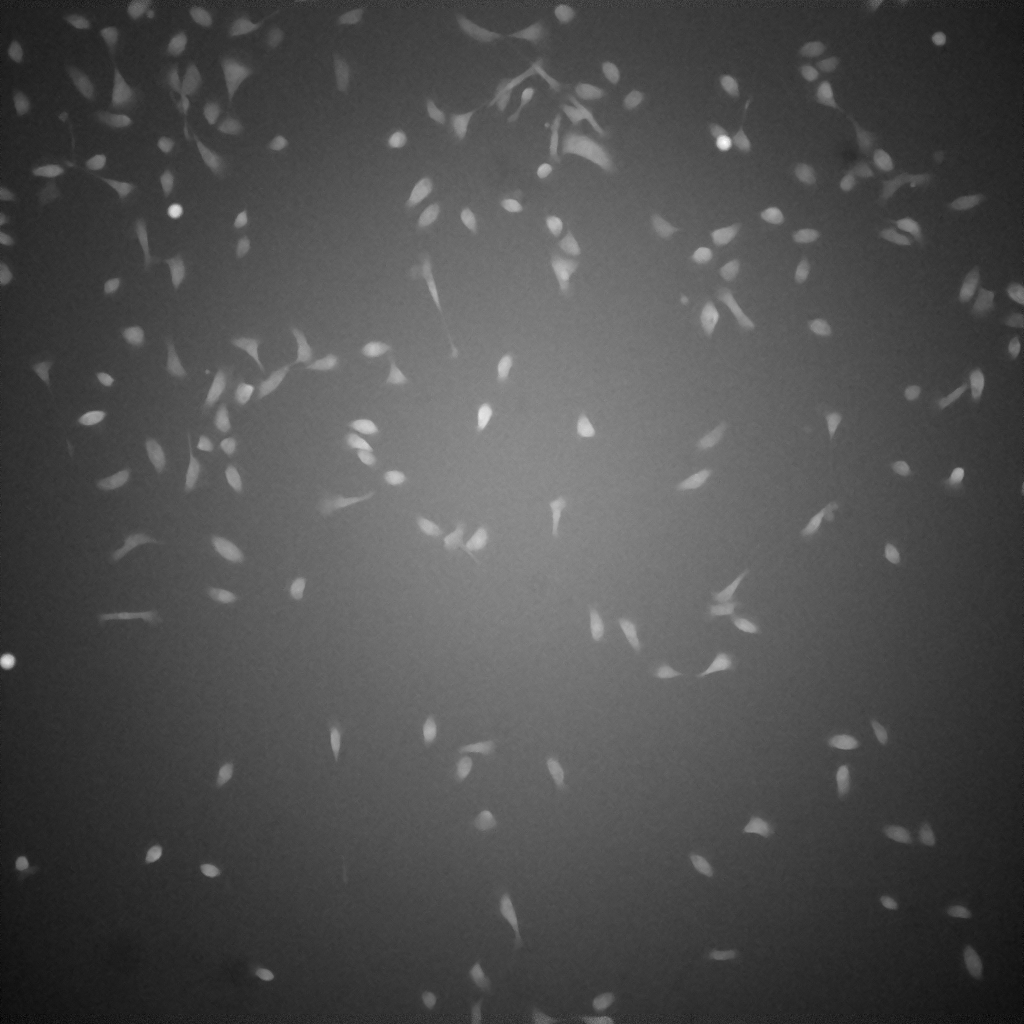}
\caption{Video of the HBEC cells, used for Figure \ref{fig:dem}C-D. To reduce the total video size, the image stack was time-subsampled 4-fold. The entire video lasts 75 hours.}
\label{si_vid:HBEC}
\end{figure}




\end{document}